\documentclass[10pt,twocolumn,letterpaper]{article}
\usepackage{cvpr}
\usepackage{times}
\usepackage{graphicx}
\usepackage{amsmath,amssymb} 
\usepackage{color}
\usepackage{algorithm}
\usepackage{algpseudocode}
\usepackage{caption}
\usepackage{subcaption}
\usepackage{multicol}
\usepackage{xspace}
\usepackage{multirow}

\newcommand{\vc}{\mathbf{c}}

\usepackage[pagebackref=true,breaklinks=true,letterpaper=true,colorlinks,bookmarks=false]{hyperref}

\cvprfinalcopy 

\def\cvprPaperID{6893} 

\ifcvprfinal\fi

\title{Multi-view 3D Reconstruction of a Texture-less Smooth Surface of Unknown Generic Reflectance}

\author{Ziang Cheng\textsuperscript{1}, Hongdong Li\textsuperscript{\rm 1}, Yuta Asano\textsuperscript{\rm 2}, Yinqiang Zheng\textsuperscript{\rm 3}, Imari Sato\textsuperscript{\rm 2}\\
\textsuperscript{\rm 1}Australian National University\\\textsuperscript{\rm 2}National Institute of Informatics,\textsuperscript{\rm 3}The University of Tokyo, Japan\\
{\tt\small \{ziang.cheng,hongdong.li\}@anu.edu.au}
}

\begin{document}

\maketitle
\begin{figure}
    \includegraphics[width=.48\textwidth]{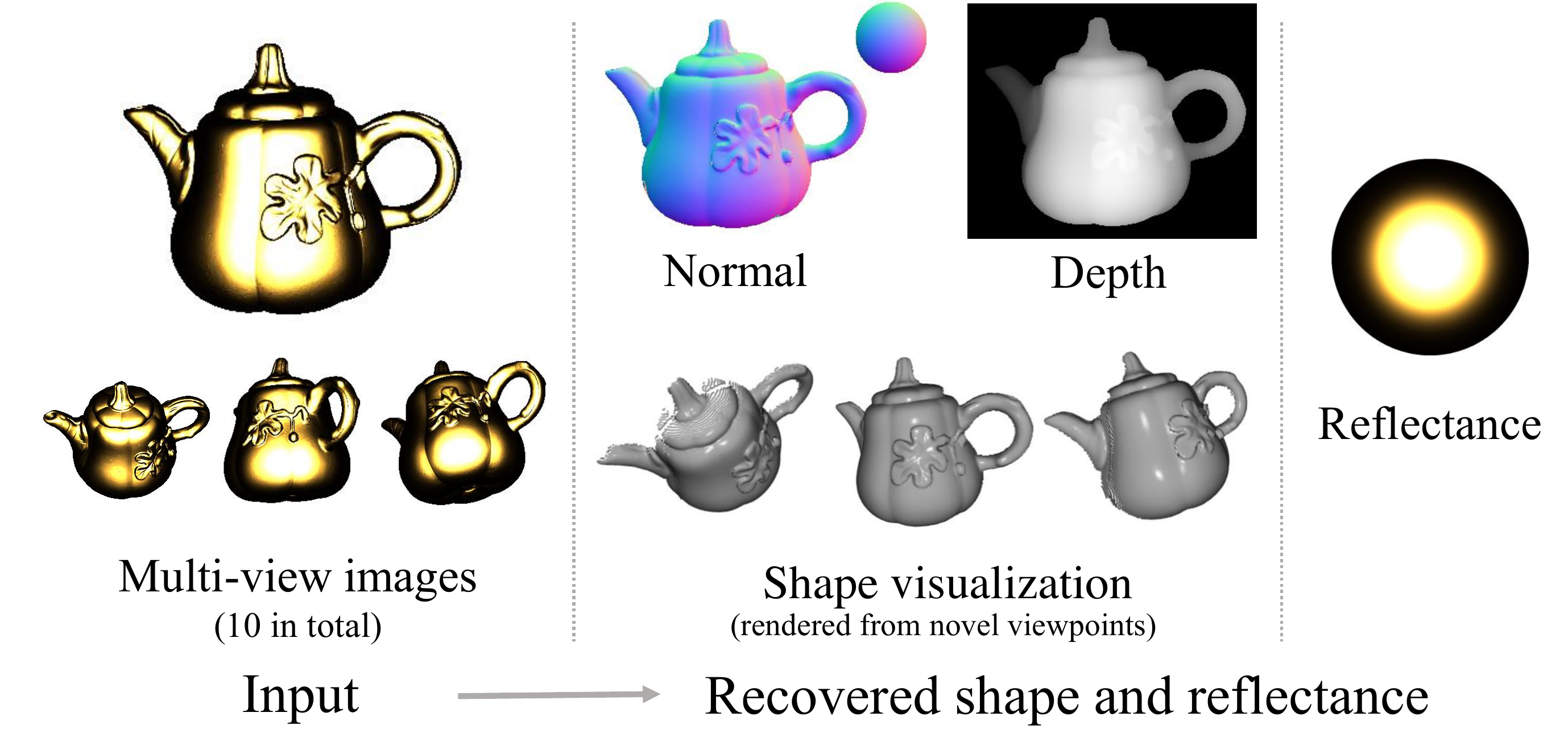}
    \captionof{figure}{\small{From a small set of input multi-view image (left), our method recovers the dense 3D object (middle) and its unknown generic surface reflectance (right).} \label{fig:my_label}}
\vspace*{-2.2em} 
\end{figure} 
\thispagestyle{empty}

\begin{abstract}
Recovering the 3D geometry of a purely texture-less object with generally unknown surface reflectance (e.g. non-Lambertian) is regarded as a challenging task in multi-view reconstruction. The major obstacle revolves around establishing cross-view correspondences where photometric constancy is violated. This paper proposes a simple and practical solution to overcome this challenge based on a {\em co-located} camera-light scanner device. Unlike existing solutions, we do not explicitly solve for correspondence. Instead, we argue the problem is generally well-posed by multi-view geometrical and photometric constraints, and can be solved from a small number of input views. We formulate the reconstruction task as a joint energy minimization over the surface geometry and reflectance. Despite this energy is highly non-convex, we develop an optimization algorithm that robustly recovers globally optimal shape and reflectance even from a random initialization. Extensive experiments on both simulated and real data have validated our method, and possible future extensions are discussed. \end{abstract}


\section{Introduction}

3D reconstruction from multi-view images is one of the central problems in computer vision. Most traditional multi-view reconstruction methods such as SFM (structure from motion) often assume the scene or object to be reconstructed have distinctive features that are view-independent, so that cross-view feature correspondences can be readily established. However, this is not the case for many commonly-seen real-world objects or surfaces manifesting non-Lambertian reflectance. Traditional SFM methods are unable to reconstruct such texture-less surfaces with glossy appearance. The problem is even more challenging if the generic surface reflectance is unknown, in which case there is no apparent way to model how object's appearance changes with viewpoint.

\begin{figure}[h!t]
    \centering
 \includegraphics[width=0.2\textwidth]{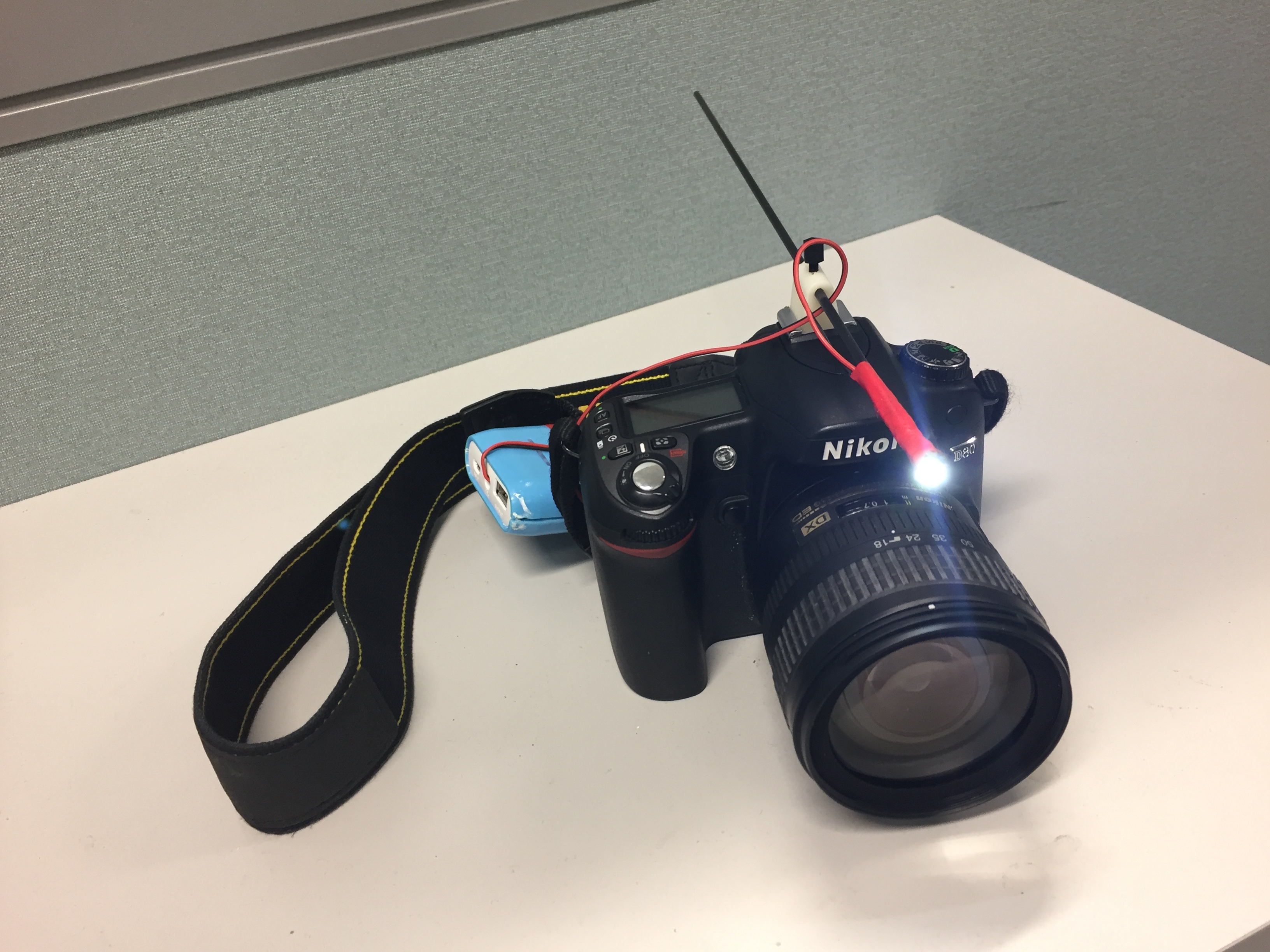}~~~\includegraphics[width=0.2\textwidth]{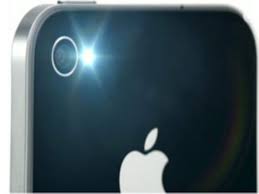}
    \caption{\small Two experiment setups: Hardware used for capturing images under a co-located setup. A point light source is rigidly attached to camera lens with a small displacement.}
    \label{fig:device}
\end{figure}

By marrying photometric stereo with traditional multi-view methods, many papers have succeeded in overcoming parts of these challenges. Most of these methods are reliant on an external initialization of the 3D shape (e.g. \cite{oxholm2012multiview,Georgoulis2014Tackling,nam2019practical}) to establish initial correspondences. Typically, finer-grained details are added incrementally to the recovered geometry. However, good initialization is not often guaranteed (\eg many require initial shape from SFM pipelines, which are already vulnerable to textureless surface or specular highlights), and a large number of input images are often required. Additionally, many methods resort to restrictive assumptions about the setup, objects and scenes. Common assumptions include \eg, purely/almost Lambertian reflectance~\cite{zhang2003shape,herandez2008mvps,higo2009handheld,wu2011fusing,beljan2012consensus,Park16robust,logothetis2019differential}, planer object shape~\cite{hui2017reflectance}, known depth via an RGB-D sensor~\cite{Schmitt_2020_CVPR}, or stereo vision with semi-static viewpoint but varying illumination~\cite{Zhou2013multiview,li2020multiview}. So far, direct multi-view reconstruction of textureless, glossy objects remains an open challenge that no method addresses well.

This paper presents a simple and practical solution to jointly recover high-fidelity surface geometry as well as unknown generic reflectance of a purely texture-less object.  
We advocate a co-located camera and light-source configuration, as shown in Fig.~\ref{fig:device}, where a point light source is rigidly and closely attached to the camera. Such scanner device is easily accessible with commodity hardware (\eg a mobile phone camera with built-in flash). 

Unlike existing photometric 3D reconstruction methods under fixed viewpoint, we allow the camera to move freely to leverage multi-view constraints. Our method is also different from existing multi-view methods mentioned above, in that we do not intend to establish explicit cross-view correspondences, either by feature matching or through shape initialization, since both are hard to obtain for a purely texture-less surface with arbitrary BRDF. Instead, we show that given a small number of views, shape and reflectance are already well-constrained by a physically-based image formation model. With this observation, we formulate reconstruction task as an energy minimization problem involving a single, unified objective. While this problem is still highly non-convex, we propose an effective optimization based approach that robustly reconstructs complex geometry as well as general reflectance without initial shape. Code and data will be available at {\small \url{https://github.com/za-cheng/PM-PMVS/}}.

\section{Related work}
{\paragraph{Multi-view Photometric stereo.}
Multi-view photometric stereo (MVPS) methods often formulate the task as energy optimization, and solve it iteratively from a coarse initialization.  The initialization is often obtained via SFM 3D reconstruction \cite{ullman1979interpretation,seitz2006comparison} or from object's visual hull \cite{laurentini1994visual}. Many methods assume Lambertian reflectance, in which case surface normal can be solved under a linear system, to which specularities are simply discarded as outliers \cite{lim2005passive,herandez2008mvps,wu2011fusing,Park16robust,logothetis2019differential}. Recovered normal field is later used to refine the 3D shape geometry, and high-frequency shape details can be gradually recovered in an iterative manner. Under known global illumination, this can be further extended to analytical BRDF models, allowing recovery of reflectance as well \cite{oxholm2012multiview}. Another notable branch of MVPS methods \cite{Zhou2013multiview,li2020multiview} uses iso-depth constraints from a light ring to propagate initial sparse correspondences from SFM. When reflectance is Lambertian, surface details can also be reconstructed by fusing shape-from-shading with classical multiview stereo \cite{langguth2016shading,kim2016multi,maurer2018combining,melou2019splitting}. However, for a textureless surface of specular material that extends outside viewing frustum, neither multi-view stereo nor visual hull is applicable, which puts a significant challenge on initializing existing MVPS methods. Additionally, MVPS methods generally require a large number of views (often a few hundreds), and most methods are designed for special scanner systems that are hard to build.

\paragraph{Co-located Photometric stereo.}
Similar to our configuration, Higo~\etal~\cite{higo2009handheld} used controlled illumination consisting of a perspective camera with a rigidly attached point light source. They assume predominantly diffusive (i.e., Lambertian) reflectance to simplify surface normal solution, and treat specularities as outliers. With a similar co-located camera-light setup, Hui~\etal~\cite{hui2017reflectance} proposed a method for general spatially-varying BRDF (SVBRDF) sampling, but it is only applicable to planar surface where pixel correspondences can be easily found (via homography).  Li \etal~\cite{li2018materials} addressed the same problem by using a single mobile phone image with the assistance of deep learning. In addition to its accessibility, a co-located setup can naturally reduce the 3D input space of isotropic BRDFs to a univariate one, which helps to improve estimation robustness even with less images. Nam~\etal~\cite{nam2019practical} later used a similar setup for joint reflectance and shape recovery. However, like most multi-view photometric methods, they require an initial shape input from a large set of images and cannot handle highly specular materials. Wang~\etal\cite{wang2020non} use a co-located light source to decouple normal and reflectance estimation in a standard photometric stereo setup. Schmitt and Donn\'e \etal~\cite{Schmitt_2020_CVPR} used a handheld RGB-D method to improve camera pose estimation, where the depth sensor is used to provide shape initialization.}

\section{Problem Setup: Image formation}\label{section_energy}

\begin{figure}[h!]
\centering
\includegraphics[width=\linewidth]{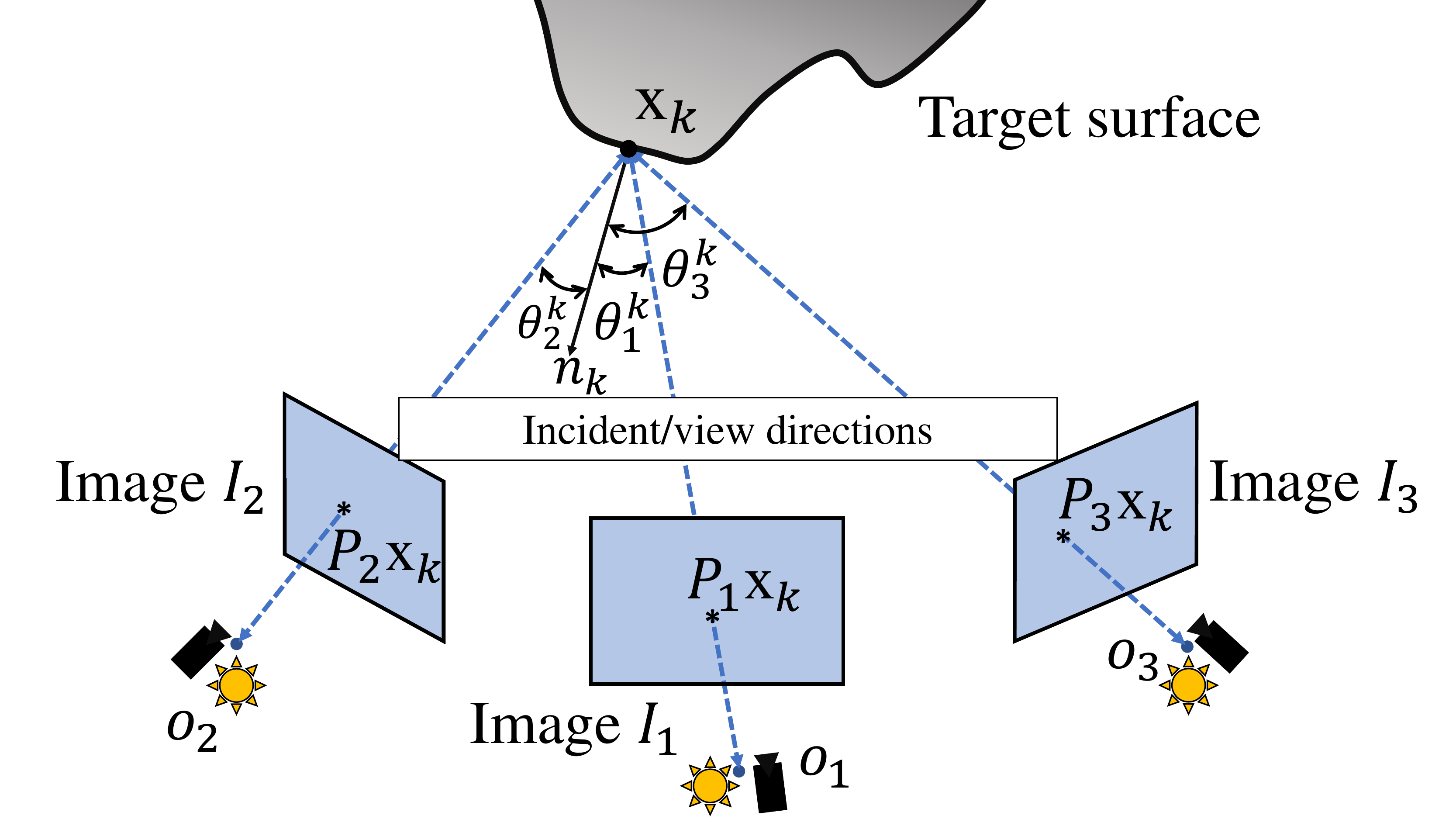}
   \caption{\small The geometric configuration of camera and light source in our reconstruction system. A light ray emitted from the point light $o_i$ hits surface point $k$. The reflected ray reaches camera, also at $o_i$, through projection $P_i$ forming multi-view observation from different viewpoints ($i$=1,2,...). Our task is to recover the surface shape (normal and depth) for all $k$, as well as reflectance.}
\label{fig:image_formation}
\vspace*{-2.2em} 
\end{figure}

The overall setup of our camera and light-source is illustrated in Fig.~\ref{fig:image_formation}, where a single point light source is rigidly attached to the camera with a small distance to the camera centre. 

\begin{figure}[h!]
\centering
\includegraphics[width=\linewidth]{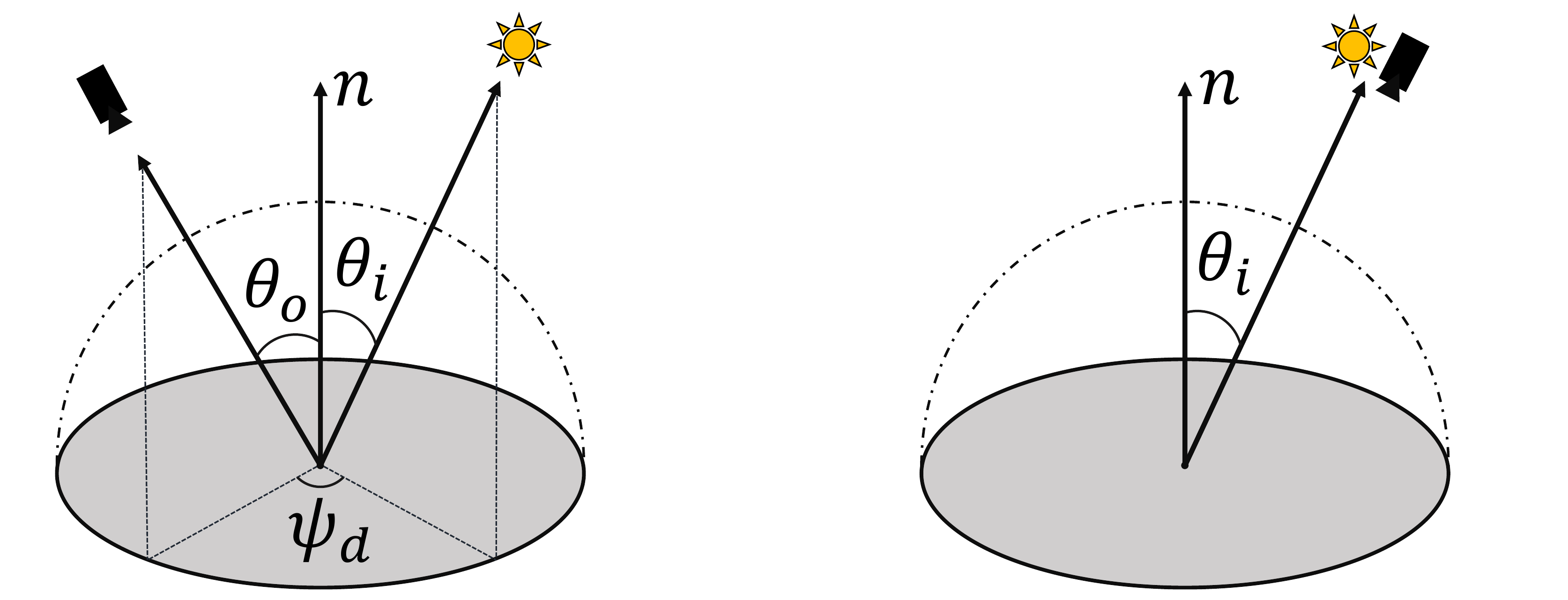}
  \caption{\small Left: Isotropic BRDF defined on three angular variables.  Right: BRDF becomes univariate function under co-located setup.}
\label{fig:brdf}
\vspace*{-2em} 
\end{figure}

We assume the object's surface BRDF is uniform and isotropic, which under out setup reduces to a univariate function of incident/view angle (Fig.~\ref{fig:brdf}). We further assume the surface is smooth (or at least piece-wise smooth) so that surface normal vectors can be defined almost everywhere.  While in this paper we mostly focus on solving spatially-uniform BRDF, our method (and its principle) can be extended to spatially-variant or {\small SVBRDF} as well. However, this is to be thoroughly addressed elsewhere to keep this paper concise and focused.  

Consider multiple images of the surface $K$ are given, each from a distinct viewpoint $m\in M$ with the above co-located system.
Denote $\mathbf{x}_k\in \mathbb{R}^3$ as the 3D world coordinates of the $k$-th surface point.  Then, its radiance observed at view $m$ is proportional to the corresponding raw pixel intensity in the $m_\text{th}$ image $I_m$.  This gives our photometric image formation equation: 
\begin{equation}
     \alpha_m^k I_m(P_m \mathbf{x}_k)={\alpha}_m^k \frac{\gamma\rho(\theta_m^k)}{d(o_m,\mathbf{x}_k)^2},
    \label{eq:linear_formation}
\end{equation} where $d(o_m,\mathbf{x}_k)=\|o_m-\mathbf{x}_k\|$ is the Euclidean distance between $o_m$ and $\mathbf{x}_k$, and $\rho(\cdot)$ is the 1D BRDF\footnote{We assume the cosine fall-off factor is subsumed in $\rho()$.}, $o_m$ is the centre of view $m$ in world coordinates, and $\theta_m^k$ denotes the incident/view angle between vector $(\mathbf{x}_k-o_m)$ and surface normal $\mathbf{n}_k$.  $P_m$ is the camera perspective projection matrix. We assume  all camera views ($P_m, m = 1...N$) are geometrically calibrated, namely they are known parameters. This is easy to achieve in practice, by using any off-the-shelf camera calibration software toolbox(\eg \cite{MatlabOTB}).         
$\gamma$ is a constant factor proportional to the brightness of light source and the response function of the camera, and $\alpha_m^k$ encodes the visibility of surface point $k$ in view $m$ (\ie, $\alpha_m^k=1$ if $k$ is visible in view $m$ and $\alpha_m^k=0$ if otherwise,\eg caused by occlusion or image boundary cropping).
  
To address the high dynamic range especially when imaging a non-Lambertian (\eg highly specular) surface, the above photometric image formation equation is often rewritten in the log-space, namely,
\begin{equation}
    \alpha_m^k \big(\log\rho(\theta_m^k) - \log I_m(P_m \mathbf{x}_k) - \log \|o_m-\mathbf{x}_k\|_2^2 + \log\gamma \big)= 0. \label{eq:formation}
\end{equation}
This log-scale form helps to improve numeric condition of the problem (ref. Nielsen~\etal \cite{nielsen2015optimal} for more details). 


\section{Energy Minimization: Joint shape and {\small BRDF} recovery} \label{section_energy}

Recall our goal is to recover the unknown  BRDF and 3D shape from multi-view observations.  We use a coefficient vector $\mathbf{c}$ to parameterize a BRDF function (\cf Sec.~\ref{brdf}). The shape is represented as a set of points $K$ indexed by pixels in the reference frame\footnote{without loss of generality, the first input image is used as the reference}. The surface shape is thus modelled as depth and normal maps in the reference frame, \ie $z_k$ and $n_k$ for all $k\in K$. Under perspective camera model, the relation between world coordinates $\mathbf{x}_k$ and depth $z_k$ can be readily established as $\mathbf{x}_k = z_k (P_\text{ref}^+p_k - o_1) + o_1$ where $P_\text{ref}^+$ is the inverse projection that maps reference frame pixel $p_k$ in image coordinates onto the unit depth plane in world coordinates. 

Our energy function is defined as a weighted sum of three energy terms: photometric term $E_p$, shape term $E_s$ and BRDF term $E_c$. 
\begin{equation}
E(\mathbf{n},\mathbf{z},\vc)= E_p(\mathbf{n},\mathbf{z},\vc) + \lambda_s E_s(\mathbf{n}, \mathbf{z}) + \lambda_c E_c(\vc) .\label{eq:energy}
\end{equation}
The first term represent the photometric constraint Eq.~\eqref{eq:formation}, and the latter two can be viewed as regularizers on shape and reflectance respectively.

\subsection{BRDF parameterization ($E_c$)}  \label{brdf}
Before we introduce our energy model, let us first define our non-parametric BRDF representation. Recent work on BRDF parameterization ~\cite{nielsen2015optimal,xu2016minimal} suggested that a wide range of real-world BRDFs can be approximated by a linear combination of a compact set of BRDF bases with high accuracy in the $\log$ space. Specifically, $\log\rho(\cdot)$ is approximated by 
\begin{equation}
   \log\rho(\cdot) \approx \texttt{D}(\cdot)\vc + \mu(\cdot) \label{BRDF}
\end{equation} 
where $\vc\in\mathbb{R}^N$ is the linear mixing coefficients, and $\mu$ is the average log-BRDF. $\texttt{D}=[d_1, d_2,..,d_N]$ is a set of pre-learned BRDF basis functions.  

Given a collection of real BRDFs (\eg MERL~\cite{Matusik:2003}), one can learn the bases as the leading $N$ eigenfunctions.  We further weight basis $d_i$ by its square-rooted eigenvalues to improve conditioning, as suggested by \cite{nielsen2015optimal}. Overall the BRDF term becomes:  
\begin{equation}
    E_c(\vc)=\|\log\rho-\texttt{D}\vc-\mu\|^2 +\|\vc\|^2 \approx \|\vc\|^2. 
\end{equation}
This term effectively encourages $\vc$ to follow a spherical Gaussian prior in our non-parametric BRDF space. A similar Tikhonov regularizer has been used to reduce the BRDF sampling as well~\cite{xu2016minimal}.

\subsection{Photometric term $E_p$}
From image formation model Eq.~\eqref{eq:formation}, a multi-view `rendering loss' of point cloud $K$ could be straightforwardly derived as 
\begin{equation}
   {\mathcal{E}}_{render}=\frac{1}{|K||M|}\sum_{k,m} \alpha_m^k L_\delta\big(\Phi_m(n_k, z_k, \vc) \big) \text{,}\label{eq:log}
\end{equation}
where $L_\delta(\cdot)$ denotes the robust Huber loss (with clipping parameter $\delta$),  $\alpha_m^k\in\{0,1\}$ is the visibility of point $k$ in view $m$ and $\Phi_m(n_k,z_k,c)$ measures the log-scale difference between expected scene radiance and true pixel intensities
{ \begin{align}
    \Phi_m(n_k,z_k,\vc) &=\texttt{D}(\theta_m^k)\vc + \mu(\theta_m^k) \nonumber\\&-\log \Big(\frac{1}{\gamma}I_m(P_m \mathbf{x}_k)\|o_m-\mathbf{x}_k\|_2^2\Big). 
\end{align} } 
A major challenge for computing Eq.(\ref{eq:log}) revolves around the visibility mask $\alpha_m^k$, which depends not only on viewpoint but also the unknown shape to be solved for, hence cannot be computed a priori. To overcome this difficulty, we use a simple heuristic to select for each surface point $k$ a subset of $\mathcal{M}$ images with the smallest photometric errors. We assume every surface points $k$ should be visible in at least $\mathcal{M}$ out of $M$ views (\ie $\forall k\in K\;\sum_{m}\alpha_m^k\ge\mathcal{M}$), where $\mathcal{M}<|M|$ is a pre-set constant. Mathematically, we define our photometric energy term as 
\begin{equation}
    E_p(\mathbf{n},\mathbf{z},\vc)=\frac{1}{|K|\mathcal{M}}\sum_{k\in K}\min_{|M_k|=\mathcal{M}}\sum_{m\in M_k} L_\delta\big(\Phi_m(n_k, z_k,\vc)\big).\label{eq:log2}
\end{equation}
Notably, the min-sum selector $M_k$, in its way of handling occluded or out-of-view points, is analogue to least trimmed squares --- an objective function commonly seen in robust regression.

\subsection{Shape term $E_s$}
A prerequisite for image formation model is the surface should be smooth, hence its normal can be defined almost everywhere and is perpendicular to tangent vectors. We design $E_s$ following this observation, to encourage surface smoothness and local integrity: 

\begin{equation}
    E_s(\mathbf{n},\mathbf{z}) = 
    \frac{1}{|K||{\mathcal{N}_k}|}\sum_k\sum_{j\in\mathcal{N}_k} \big(n_k^T(\mathbf{x}_k-\mathbf{x}_j)\big)^2
    , \label{E_s}
\end{equation}
where $\mathcal{N}_k$ is the 4-neighbor of pixel $k$ in the reference frame.

\section{Solving the energy minimization} \label{solution}
The above energy function is highly non-convex due to the view-dependent non-Lambertian BRDF and the arbitrary image measurements in the scene. Previous photometric methods tackled this non-convexity either by using overly simplistic reflectance model (\eg pure Lambertian) or by assuming the availability of high quality initialization, if not both.  

In this paper, we do not rely on above assumptions. Our energy minimization algorithm is based on {\em coordinate descent}, alternating between two sub-problems of solving BRDF and solving shape parameters respectively (see Sec.~\ref{sec:solve_brdf} and Sec.~\ref{sec:solve_shape}). Therefore it can be initialized from an inexact estimation of either shape or BRDF. In fact, we show it can converge quickly and correctly even from a {\em null} initialization, namely, starting from the triviality of $\vc=0$ with no initial shape.

\subsection{Solve for BRDF, fixing Shape}\label{sec:solve_brdf}
Suppose a current 3D shape estimation is given and fixed, we solve for the BRDF by minimizing Eq.~\eqref{eq:energy}, \ie,  
\begin{equation}
    \min_{\vc,\gamma} E_p(\mathbf{n},\mathbf{z},\vc) + \lambda_c E_c(\vc). \label{min_c}
\end{equation}
Here we keep the minimum set $M_k$ constant during the optimization, in which case $E_p(\mathbf{n},\mathbf{z},\cdot)$ becomes convex and can be globally minimized. We note that in practice such approximation has minimal impact on the estimation due to problem being well-constrained on $\vc$, and the approximation is indeed an upper bound of true energy. 
We employ a standard L-BFGS optimizer \cite{nocedal1980updating} and solve the camera response rate $\gamma$ together with $\vc$. 

\subsection{Solve for Shape, fixing BRDF}\label{sec:solve_shape}
With fixed BRDF, surface shape is solved by minimizing Eq.~\eqref{eq:energy} \wrt $\mathbf{n},\mathbf{z}$. 
This is a highly challenging minimization problem due to the non-convexity of the image formation function $I_m(P_m \cdot)$ and BRDF $\rho_c(\cdot)$.  In our experiments we found that conventional optimization algorithms (\eg gradient-descent or quasi-Newton) almost always fail unless provided with a high quality initialization. 

Formally, we seek to solve the following minimization problem: 
\begin{equation}
    \min_{\mathbf{n},\mathbf{z}} E_p(\mathbf{n},\mathbf{z},\vc) + \lambda_s E_s(\mathbf{n},\mathbf{z}). \label{solve_shape}
\end{equation}
To do this, we introduce a set of auxiliary variables $\tilde{\mathbf{z}}=[\tilde{z}_1,...,\tilde{z}_K]^T$ to decouple $E_p$ and $E_s$, and employ a quadratic penalty method ({\small QPM}) \cite{steinbrucker2009large} to relax the `hard' constraint $\tilde{\mathbf{z}} - \mathbf{z}=0$
\begin{equation}
E_{\text{QPM}}(\mathbf{n},\mathbf{z},\tilde{\mathbf{z}}) = E_p(\mathbf{n},\mathbf{z},\vc) + \lambda_sE_s(\mathbf{n}, \tilde{\mathbf{z}}) + \sigma^{(i)} \|\tilde{\mathbf{z}} - \mathbf{z}\|^2 \label{penalty_optimization},
\end{equation}
where $\sigma^{(i)}=\kappa\sigma^{(i-1)}$ is a penalty coefficient that increases exponentially \wrt $i$ by some factor $\kappa>1$. As $i$ grows, violations of $\tilde{\mathbf{z}}-\mathbf{z}=0$ are penalized with increased severity until the constraint is satisfied. In this paper we use fixed $\kappa=1.3$.

The purpose for above relaxation is that unlike Eq.(\ref{solve_shape}), Eq.(\ref{penalty_optimization}) is now convex (quadratic) \wrt $\tilde{\mathbf{z}}$, and $k$-\textit{separable} \wrt $\mathbf{n}, \mathbf{z}$. In other words, the objective \wrt $\mathbf{n}, \mathbf{z}$ can be re-written as a summation over $K$ mutually independent sub-energy terms \footnote{See supplemental for proof.}
\begin{equation}
    E_{\text{QPM}}(\mathbf{n},\mathbf{z},\tilde{\mathbf{z}})=\sum_{k\in K}E_{\text{QPM}}^k(\mathbf{n}_k,z_k,\tilde{\mathbf{z}}),
\end{equation}
where each term $E_{\text{QPM}}^k(\cdot,\cdot,\tilde{\mathbf{z}})$ has a small input space that can be directly searched. Therefore $E_{\text{QPM}}(\cdot,\cdot,\tilde{\mathbf{z}})$ as a whole can be globally minimized by \eg exhaustive search within complexity $O(K)$\footnote{Global optimality is guaranteed under the weak assumption that the depth $z_k$ is bounded and energy is bounded Lipschitz continuous, see supplemental for proof.}. This is a crucial step as it allows one to overcome the non-convexity of original energy (most of which is in $\mathbf{n}, \mathbf{z}$ dimensions) in a tractable manner.

We solve the QPM optimization by alternating coordinate descent between $\{\mathbf{n},\mathbf{z}\}$ and $\tilde{\mathbf{z}}$, as listed in Alg.~\ref{algorithm}. Similar to the arguments made above, the optimal $\tilde{\mathbf{z}}$ is amenable to closed-form solution by the sparse least square {\small LSQR} algorithm~\cite{lsqr}. Conversely, optimization of $\{\mathbf{n},\mathbf{z}\}$ is non-convex, but can be decomposed into $K$ independent sub-problems.

Here we solve the latter using a randomized search approach inspired by PatchMatch~\cite{barnes2009patchmatch,barnes2010generalized}, which iterates between two steps: \textit{propagation} and \textit{randomized search}. During \textit{propagation} step, for every $k$ we attempt to improve $E_{\text{QPM}}^k$ using its neighbors' normal and depth $\{\mathbf{n}_j,z_j|j\in \mathcal{N}_k\}$ as candidates. During the \textit{randomized search}, we attempt to improve $E_{\text{QPM}}^k$ with random candidates of $\mathbf{n}_k,z_k$. In both steps, the best candidate is kept and carried to the next iteration. In practice, we found PatchMatch to be significantly more efficient than exhaustive search, thanks to the spatial smoothness of shape variables. An optimal solution can often be found in just 10 to 15 iterations. Note that PatchMatch does not need an initialization. Rather it starts from a random initial point. We refer the readers to Barnes \etal \cite{barnes2009patchmatch,barnes2010generalized} for details about PatchMatch.
{\small 
\begin{algorithm}[!htb]
\begin{algorithmic}
\Function{SolveShape}{$\mathbf{n},\mathbf{z},\sigma^{(0)}$}
\State $\tilde{\mathbf{z}} = \mathbf{0}$
\State $\sigma=\sigma^{(0)}$ \Comment{initially very small} 
\Repeat
    \State {$\mathbf{n},\mathbf{z}$ = \Call{PatchMatch}{min\_func=$E_{\text{QPM}}(\cdot,\cdot,\tilde{\mathbf{z}})$}}
    \State {$\tilde{\mathbf{z}}$ = \Call{lsqr}{min\_func=$E_{\text{QPM}}(\mathbf{n},\mathbf{z},\cdot)$}}
    \If {$\|\mathbf{z}-\tilde{\mathbf{z}}\|$ is not sufficiently small}
        \State $\sigma=\kappa \sigma$
        \State{\textbf{continue}}
    \EndIf
\Until{Energy converges}
\State \Return $\mathbf{n},\mathbf{z}$
\EndFunction
\end{algorithmic}
\caption{\small Solution for Eq.~\eqref{solve_shape} by solving a series of quadratic coupling sub-problems.}\label{algorithm}
\end{algorithm}
}

An overview of relaxed minimization is presented in Alg.~\ref{algorithm}. Note the solution at least converges to a local minimum, as eventually neither PatchMatch nor LSQR increases energy when $\sigma$ no longer changes. In practice, we found the solution turned out to be almost always globally optimal.

\subsection{Analysis} 
\begin{figure}[!htb]
\centering
    \begin{subfigure}[b]{0.22\textwidth}
    \includegraphics[width=\textwidth, trim={1cm 1cm 0cm 1cm}]{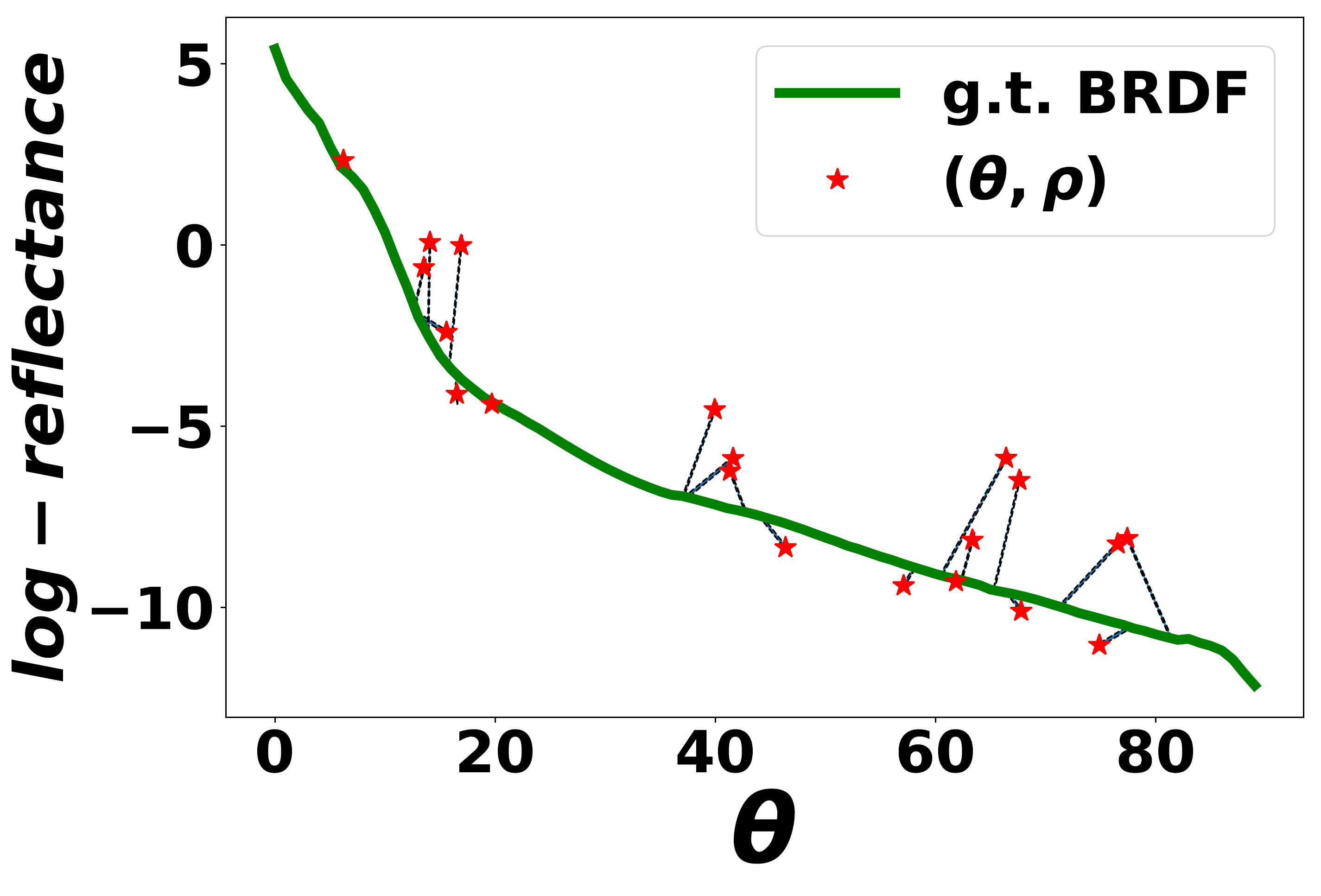}
    \caption{ with wrong $\mathbf{x}_k,n_k$\label{fig:scatterc}}
    \end{subfigure}
    ~~~
    \begin{subfigure}[b]{0.22\textwidth}
    \includegraphics[width=\textwidth, trim={1cm 1cm 0cm 1cm}]{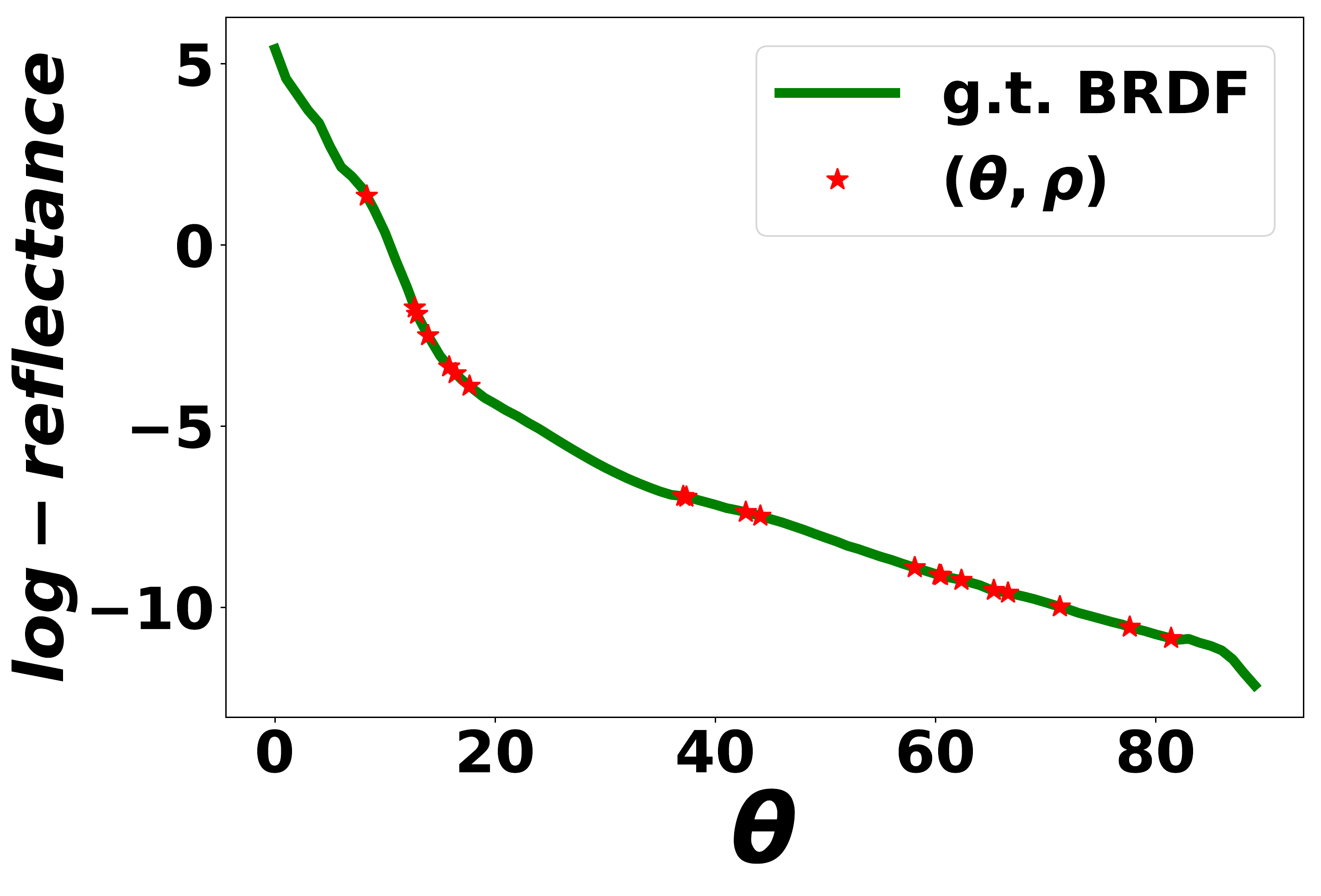}
    \caption{with ground-truth shape\label{fig:scattera}}
    \end{subfigure}
    \caption{\small Shape and BRDF visualized in $\theta$-reflectance domain. The green curve is the ground truth BRDF and red points are inferred from shape under Eq.~\eqref{eq:formation}. The latter is nonconforming to any BRDF if shape is given wrong values (\ref{fig:scatterc}). On the other hand, the ground-truth shape corresponds to a true minimizer as shown in (\ref{fig:scattera}).}
    \label{fig:scatter}
    \vspace*{-2.2em} 
\end{figure}

Before presenting our experiment results,let us analyze the existence of optimal solution. Specifically, we will show that ground-truth (shape and BRDF) is indeed a valid optimizer for the energy function, and conversely, minimizing this energy to its optimum will lead to the true solution.  We do so by considering two reflectance estimations: one is the ground-truth BRDF itself, and the other is inferred from the estimated shape $\{z_k,\mathbf{n}_k\}$ by solving Eq.~\eqref{eq:formation}, as illustrated in Fig.~\ref{fig:scatter}. The equality of Eq.~\eqref{eq:formation} states that the two estimations must agree with each other. Otherwise a discrepancy generally exists between the two, which contributes to a non-minimum energy value. By minimizing the energy, such discrepancy is also reduced, eventually leading to the true BRDF and true shape.  (The reader is also referred to the curves obtained in our real experiments in Fig~.\ref{fig:visual_constraint} for a better visualization.) 

\begin{figure*}[!htbp]
    \centering
    \captionsetup[figure]{font=tiny}
    \begin{subfigure}[b]{0.15\textwidth}
        \centering
        \includegraphics[width=\textwidth,trim={6cm 1cm 3.5cm 1.5cm},clip]{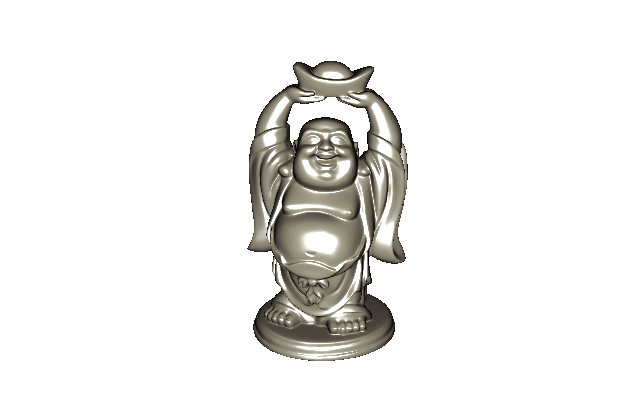}
        \includegraphics[width=\textwidth,trim={6cm 3cm 3.5cm 1.5cm},clip]{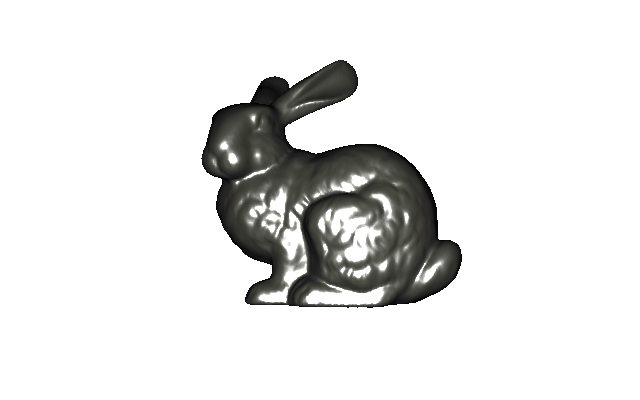}
        \includegraphics[width=\textwidth,trim={5cm 3cm 4.5cm 1.5cm},clip]{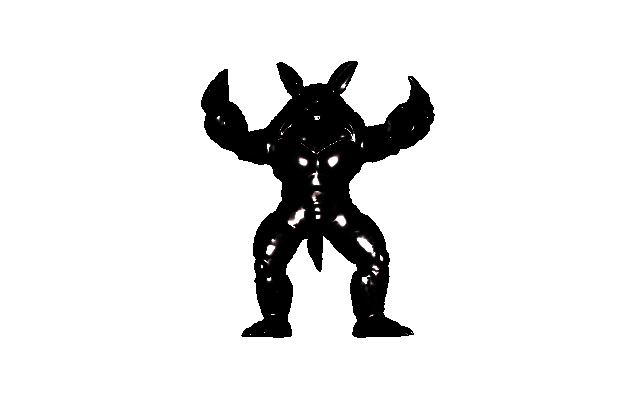}
        \includegraphics[width=\textwidth,trim={5cm 3cm 4.5cm 1.5cm},clip]{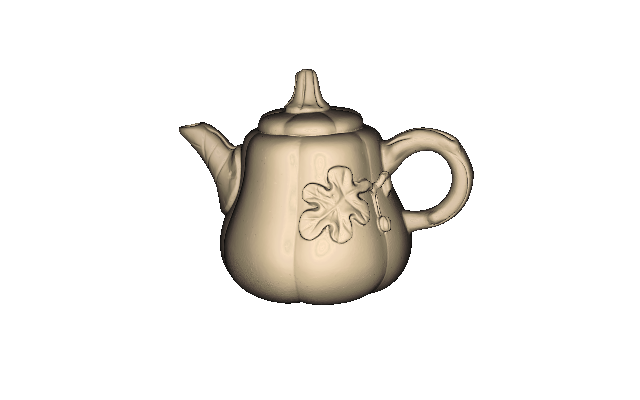}
        \caption{\scriptsize ref. view}
    \end{subfigure}
    \begin{subfigure}[b]{0.15\textwidth}
        \centering
        \includegraphics[width=\textwidth,trim={6cm 1cm 3.5cm 1.5cm},clip]{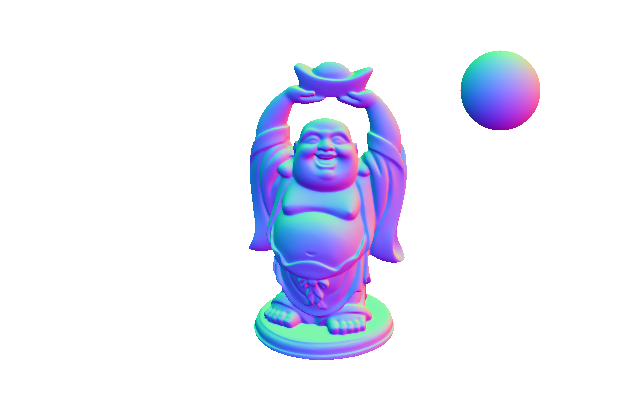}
        \includegraphics[width=\textwidth,trim={6cm 3cm 3.5cm 1.5cm},clip]{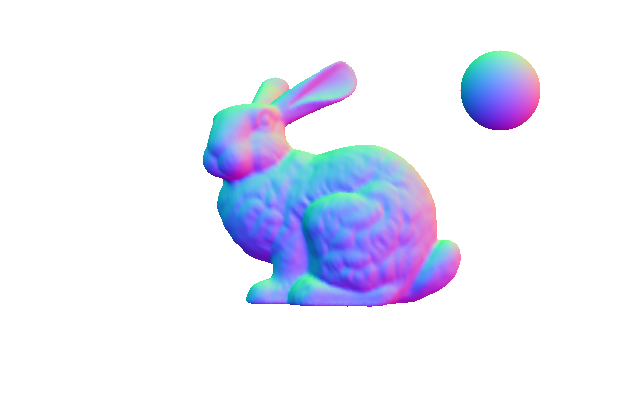}
        \includegraphics[width=\textwidth,trim={6cm 3cm 3.5cm 1.5cm},clip]{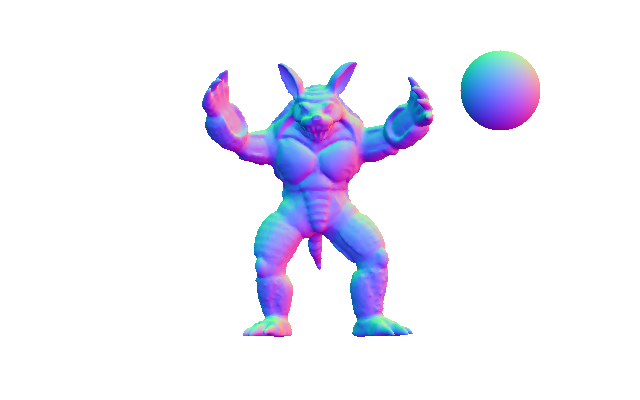}
        \includegraphics[width=\textwidth,trim={6cm 3cm 3.5cm 1.5cm},clip]{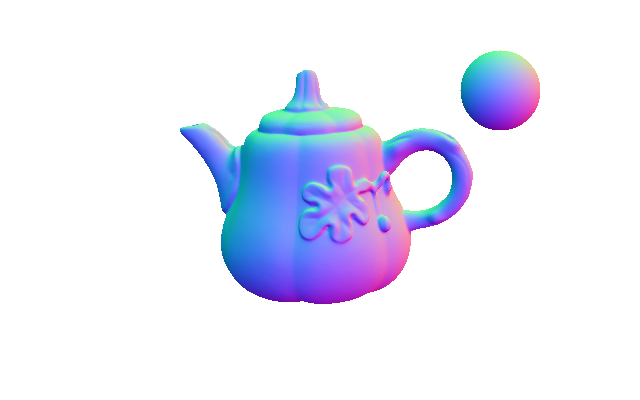}
        \caption{\scriptsize g.t. normal}
    \end{subfigure}
    \begin{subfigure}[b]{0.15\textwidth}
        \centering
        \includegraphics[width=\textwidth,trim={5cm 1cm 4.5cm 1.5cm},clip]{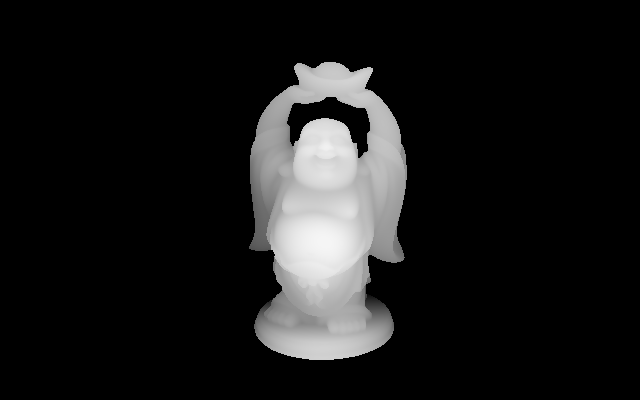}
        \includegraphics[width=\textwidth,trim={5cm 3cm 4.5cm 1.5cm},clip]{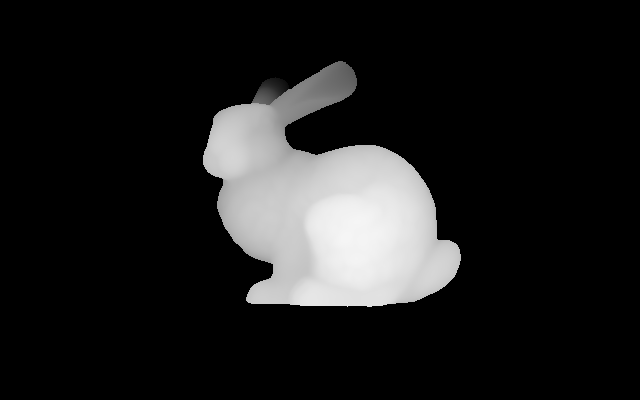}
        \includegraphics[width=\textwidth,trim={5cm 3cm 4.5cm 1.5cm},clip]{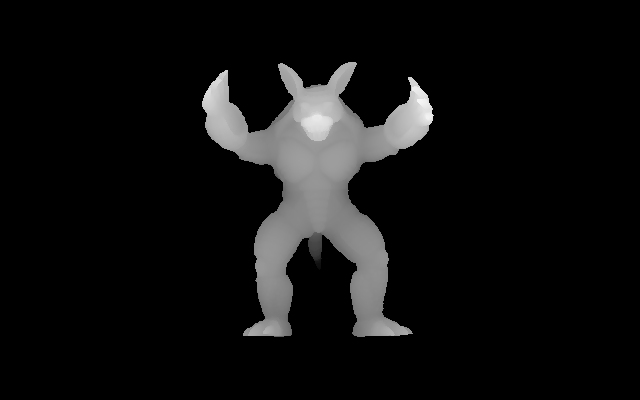}
        \includegraphics[width=\textwidth,trim={5cm 3cm 4.5cm 1.5cm},clip]{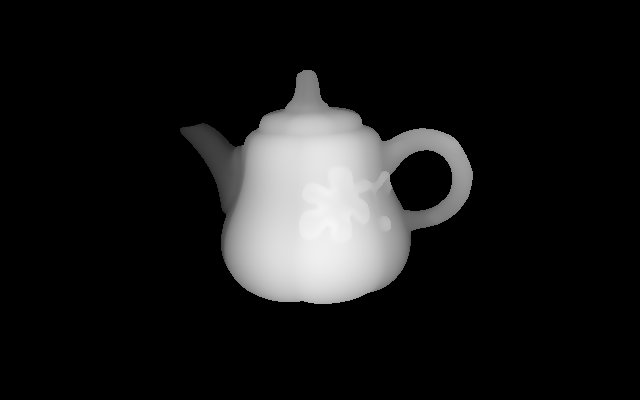}
        \caption{\scriptsize g.t. depth}
    \end{subfigure}
    \begin{subfigure}[b]{0.15\textwidth}
        \centering
        \includegraphics[width=\textwidth,trim={6cm 1cm 3.5cm 1.5cm},clip]{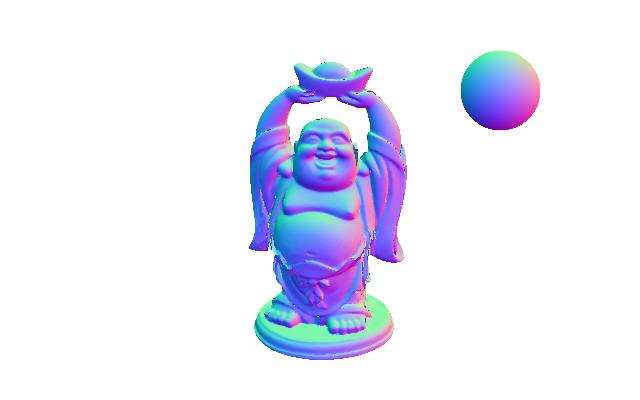}
        \includegraphics[width=\textwidth,trim={6cm 3cm 3.5cm 1.5cm},clip]{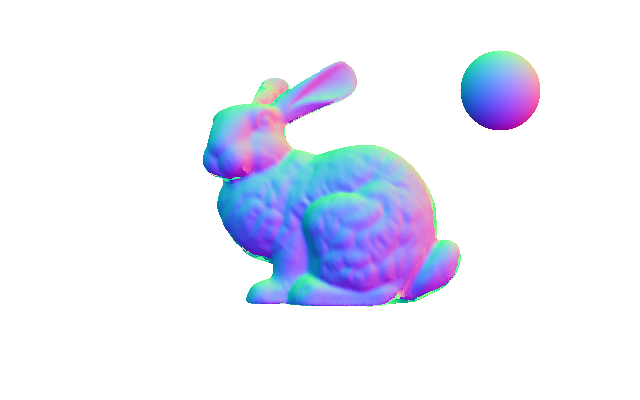}
        \includegraphics[width=\textwidth,trim={6cm 3cm 3.5cm 1.5cm},clip]{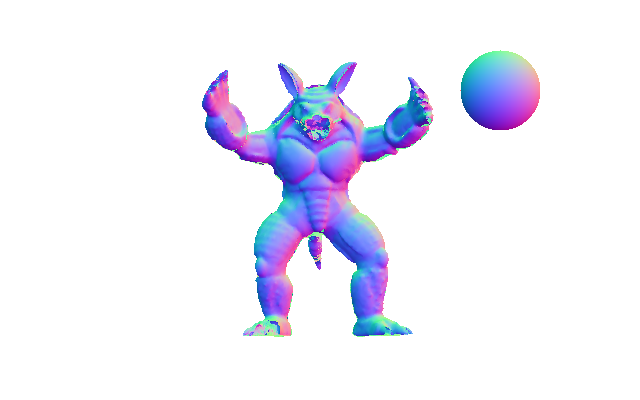}
        \includegraphics[width=\textwidth,trim={6cm 3cm 3.5cm 1.5cm},clip]{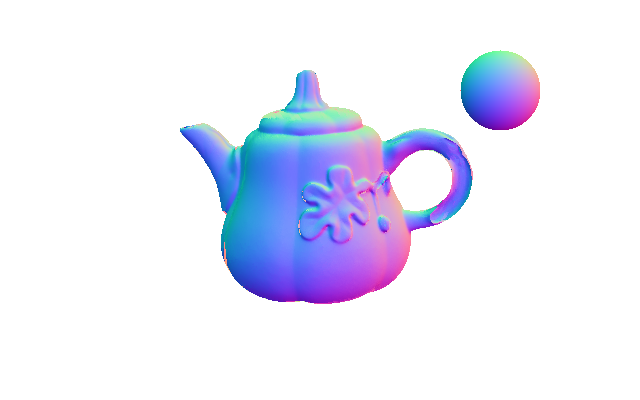}
        \caption{\scriptsize est. normal}
    \end{subfigure}
    \begin{subfigure}[b]{0.15\textwidth}
        \centering
        \includegraphics[width=\textwidth,trim={5cm 1cm 4.5cm 1.5cm},clip]{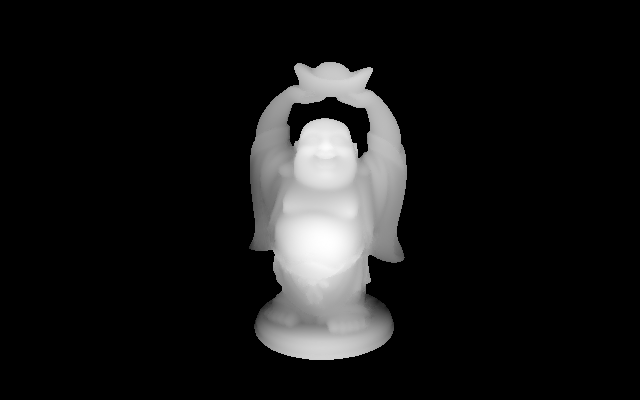}
        \includegraphics[width=\textwidth,trim={5cm 3cm 4.5cm 1.5cm},clip]{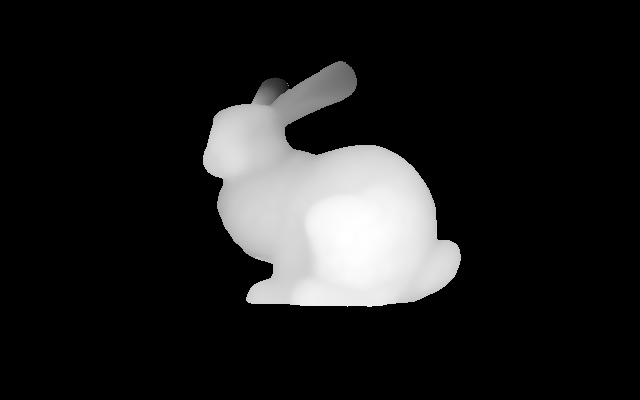}
        \includegraphics[width=\textwidth,trim={5cm 3cm 4.5cm 1.5cm},clip]{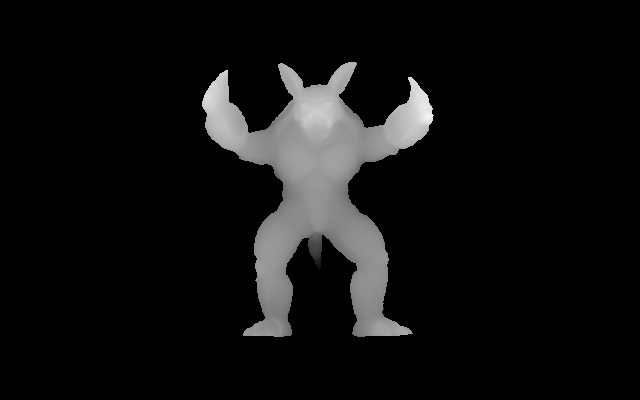}
        \includegraphics[width=\textwidth,trim={5cm 3cm 4.5cm 1.5cm},clip]{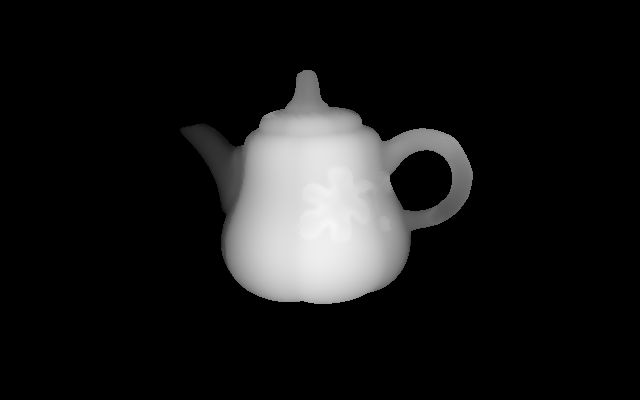}
        \caption{\scriptsize est. depth}
    \end{subfigure}
    \begin{subfigure}[b]{0.15\textwidth}
        \centering
        \includegraphics[width=\textwidth,trim={6cm 1cm 3.5cm 1.5cm},clip]{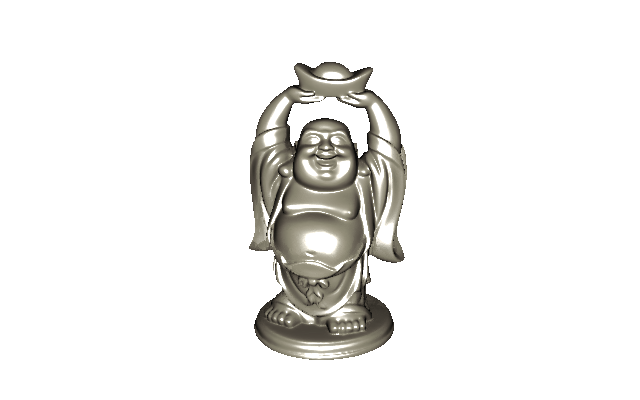}
        \includegraphics[width=\textwidth,trim={6cm 3cm 3.5cm 1.5cm},clip]{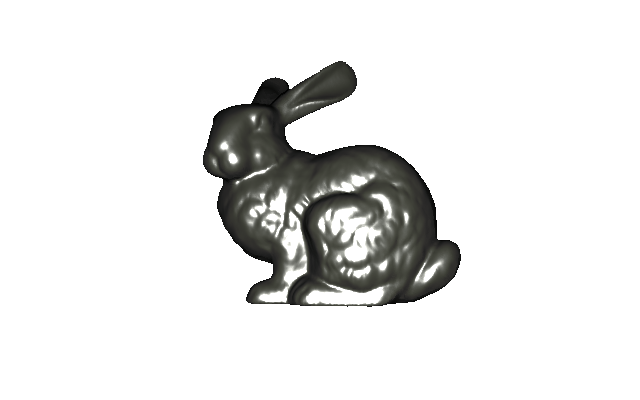}
        \includegraphics[width=\textwidth,trim={5cm 3cm 4.5cm 1.5cm},clip]{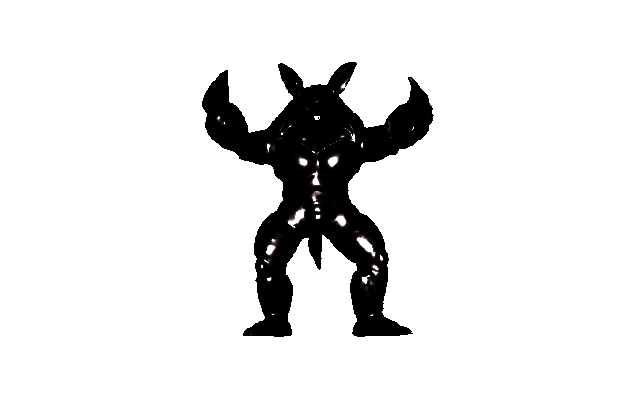}
        \includegraphics[width=\textwidth,trim={5cm 3cm 4.5cm 1.5cm},clip]{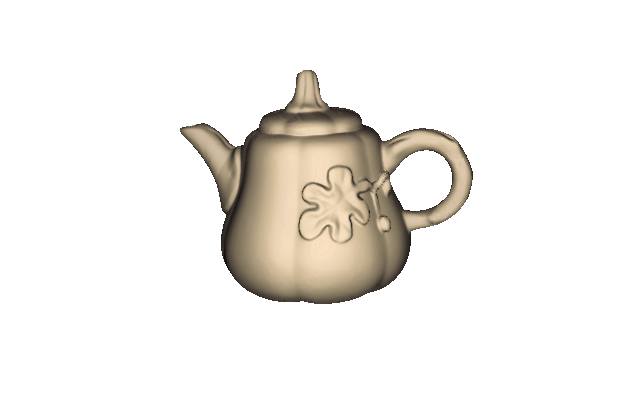}
        \caption{\scriptsize re-render}
    \end{subfigure}
    \caption{\small Results on Synthetic objects. See supplemental for more visual results.}
    \label{fig:vis_results}
    \vspace*{-1.5em} 
\end{figure*}

\begin{figure}
    \centering
    \includegraphics[width=0.48\textwidth]{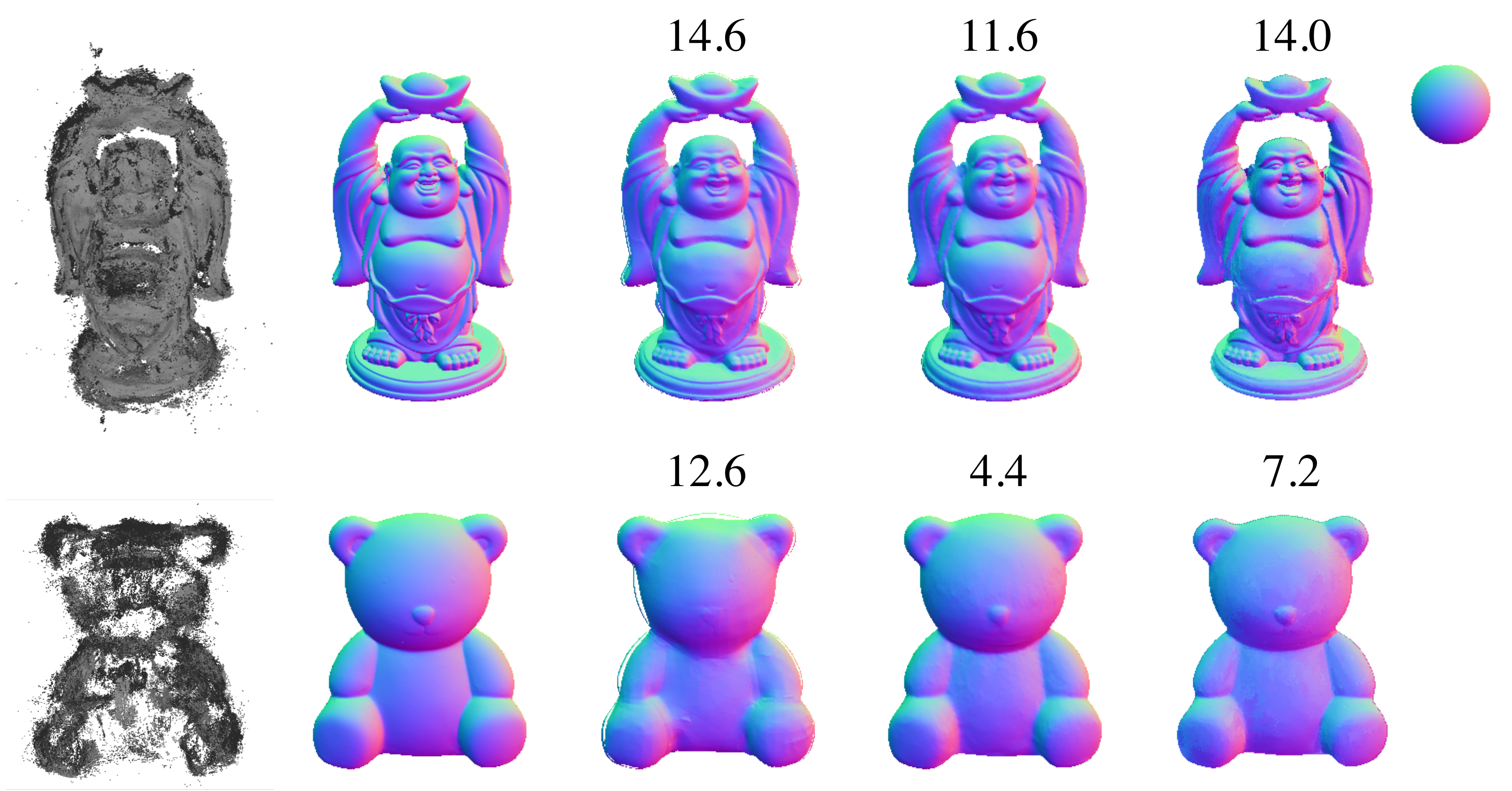}
    
    \begin{subfigure}[b]{0.085\textwidth}
    \centering\caption*{\scriptsize \cite{schoenberger2016sfm}}
    \end{subfigure}
    \begin{subfigure}[b]{0.085\textwidth}
    \centering\caption*{\scriptsize G.T.}
    \end{subfigure}
    \begin{subfigure}[b]{0.085\textwidth}
    \centering\caption*{\scriptsize \cite{Park16robust}}
    \end{subfigure}
    \begin{subfigure}[b]{0.085\textwidth}
    \centering\caption*{\scriptsize \cite{li2020multiview}}
    \end{subfigure}
    \begin{subfigure}[b]{0.10\textwidth}
    \centering\caption*{\scriptsize Ours}
    \end{subfigure}
    \begin{subfigure}[b]{0.005\textwidth}
    \centering\caption*{\scriptsize }
    \end{subfigure}
    \vspace*{-1em} 
    \caption{Comparison on Diligent-MV dataset \cite{li2020multiview} with normal errors listed on top. Note our method received far less images than the other two methods, and does not require high quality initial shape.}
    \label{fig:diligent_compare}
    \vspace*{-1.5em} 
\end{figure}

\section{Experiments}\label{experiment}
To validate the proposed method, we conducted experiments on both synthetic and real images of multi-view photometric observations. 
\subsection{Synthetic experiments}
For synthetic experiments, our aim is to quantitatively verify the effectiveness of our method under different imaging conditions.  Throughout our experiments, we use the same and fixed set of parameter settings, \ie, $N=15$, $\delta=0.1$, $\lambda_s=10^6$, $\lambda_c=0.005$, suggesting the method is rather agnostic or robust to meta-parameters.  Our energy minimization algorithm was always initialized from null BRDF \ie $\mathbf{c=0}$ and without initial shape.

We render multiple 3D mesh models with BRDFs sampled form the MERL database~\cite{Matusik:2003}.  To validate the BRDF models, we perform multiple rounds of cross validation where 95 out of the 100 materials in MERL were used to learn the bases (or dictionary) $\texttt{D}$ and the remaining five BRDFs for rendering and testing. We position the virtual camera approximately one world unit (metre) away from mesh in reference frame, and all objects (except the open surface) are scaled so that they span 0.25 unit length (25 centimetres) in whichever is the greatest of its X, Y and Z dimensions. We render only 10 views of the target objects with co-located point light source.

Some reconstructions obtained from our method are visualized in Fig~\ref{fig:vis_results}, where we test the performance on four models `buddha, bunny, armadillo, teapot' from Stanford dataset~\cite{Curless1996AVM} and Diligent dataset~\cite{Shi2019ABD}. Additionally, we render an open and smooth surface `\textit{himmelblau}' (refer to supplemental for details). In table-\ref{tab:errors} we list quantitative performance \wrt ground truth shape and BRDF.  We measure the mean errors for recovered surface normal in degrees, depth map (in world units), and log-BRDF (absolute difference averaged over input angles in $[0,\pi/2)$). Note the mean normal and depth errors are mostly caused by heavy occlusion on objects' boundaries where the shape cannot be exactly recovered, while the median errors are mush smaller.

\subsection{Comparison}
To the best of our knowledge, our method is original and few previous paper had attempted at this challenging task in the same setting. This makes a direct and fair comparison difficult. However, to provide the reader a sense of the performance of our 3D reconstruction, we offer comparison with two state-of-the-art methods -- Park \etal \cite{Park16robust} and Li \etal \cite{li2020multiview} -- on Diligent-MV dataset \cite{li2020multiview}. 
We note that this requires us to modify our method for a detachable light source (\ie a non-co-located setup), and model a higher dimensional BRDF. In this experiment, we follow the bi-variate BRDF approximation, and incorporate a second difference angle in our BRDF formulation \cite{rusinkiewicz1998new}. 

Furthermore, we note following differences between our methods and \cite{Park16robust,li2020multiview}: (1) our method is based on 21 input images from 3 viewpoints, while \cite{Park16robust,li2020multiview} received 1920 images from 20 viewpoints (2) we reconstruct an oriented point cloud indexed by reference pixels, while \cite{Park16robust,li2020multiview} outputs a water tight triangle mesh (3) our method is randomly initialized, while \cite{Park16robust,li2020multiview} received high quality initial shape from MVS pipeline \cite{galliani2015massively} with human correspondence labeling involved.

Fig \ref{fig:diligent_compare} illustrates the reconstructed normal maps on two real world models \textit{Buddha} and \textit{Bear}, with corresponding mean normal error listed on top. We also include the results from COLMAP \cite{schoenberger2016sfm} as an SFM baseline. We are able to achieve comparable performance to \cite{Park16robust,li2020multiview} despite using far less images and unaided by initial geometry. Compared to COLMAP \cite{schoenberger2016sfm}, we reconstruct a dense point cloud of arguably better quality.

\begin{table}[!htb]
    \centering
    \scriptsize
    \caption{\small Error metrics on different target objects. Note we erode foreground region by 2 pixels for evaluation purpose since object's boundaries are often heavily self-occluded.}
    \label{tab:errors}
    \begin{tabular}{c*{6}{|c}}
         \multirow{2}{*}{Models} & \multicolumn{2}{|c}{Normal in degrees} & \multicolumn{2}{|c}{Depth $\times$ 1000} & \multicolumn{2}{|c}{BRDF $\times$ 10}\\\cline{2-7}
         & median & mean& median & mean& median & mean\\
         \hline
          Buddha & 1.87& 5.59    &0.45 & 1.36  & 0.37 & 1.44
         \\\hline
          Bunny & 1.33& 5.27   & 0.42& 0.90   & 0.30& 2.81
         \\\hline
          Armadillo & 1.40& 7.81   &0.35 & 1.57   &0.66 & 1.33
         \\\hline
          Teapot & 0.67& 2.64  &0.51 & 0.82   & 0.38& 0.55
         \\\hline
          Himmelblau & 1.45& 2.06  & 1.28& 5.40   &0.67 & 1.94
         \\\hline
          \textbf{Overall} &1.36 & 5.65  &0.59 & 1.40   & 0.69& 1.56
          \\\hline
    \end{tabular}
\end{table}

\begin{figure}[htb]
    \centering
    \hspace*{-0.5cm}
    \begin{subfigure}[b]{0.18\textwidth}
         \centering
         \includegraphics[trim={0 0 10 0},clip,height=0.7\textwidth]{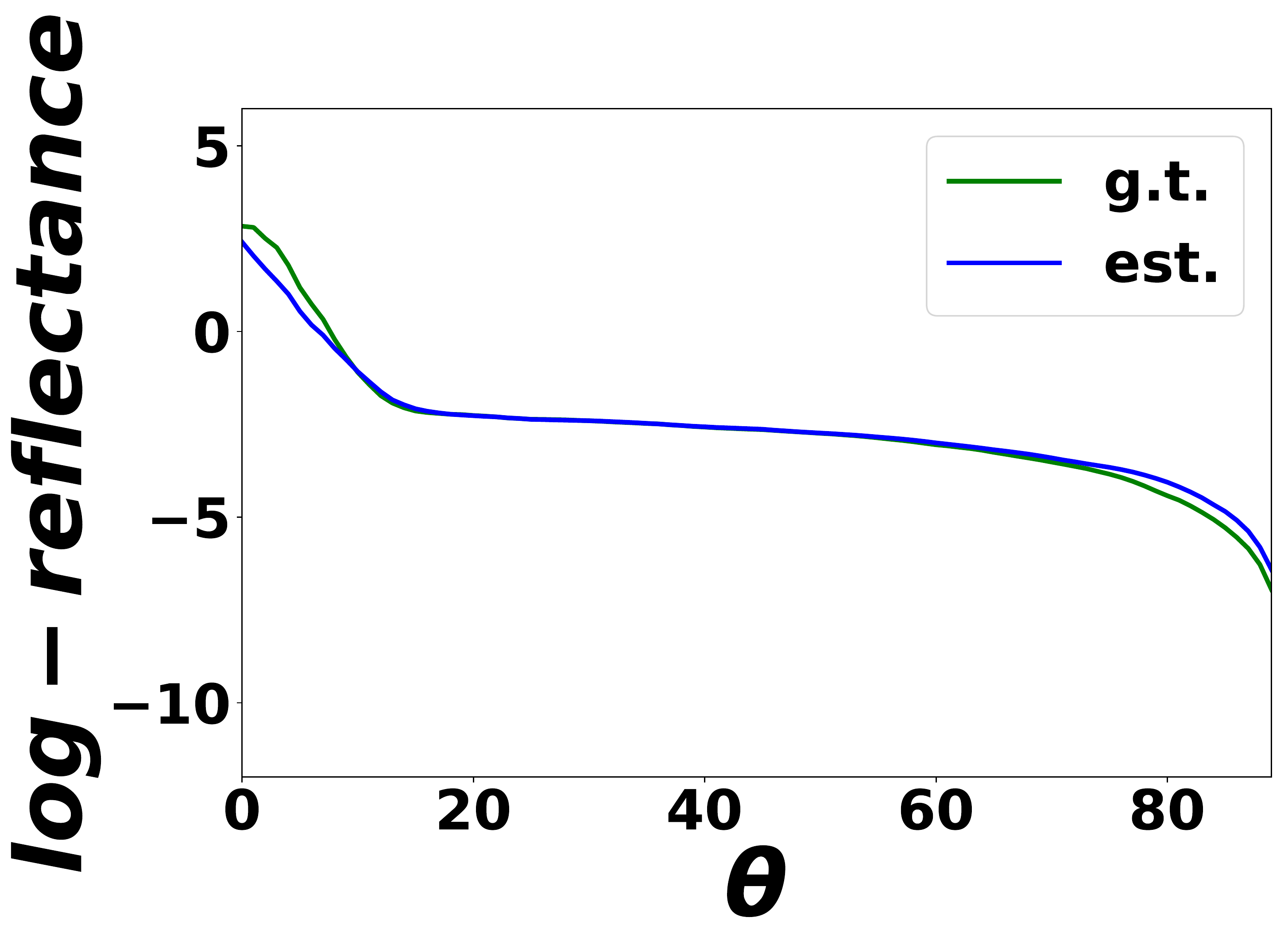}
         \caption{\footnotesize alumina-oxide}
     \end{subfigure}
    \hspace*{-0.2cm}
    \begin{subfigure}[b]{0.18\textwidth}
         \includegraphics[trim={70 0 10 0},clip,height=0.7\textwidth]{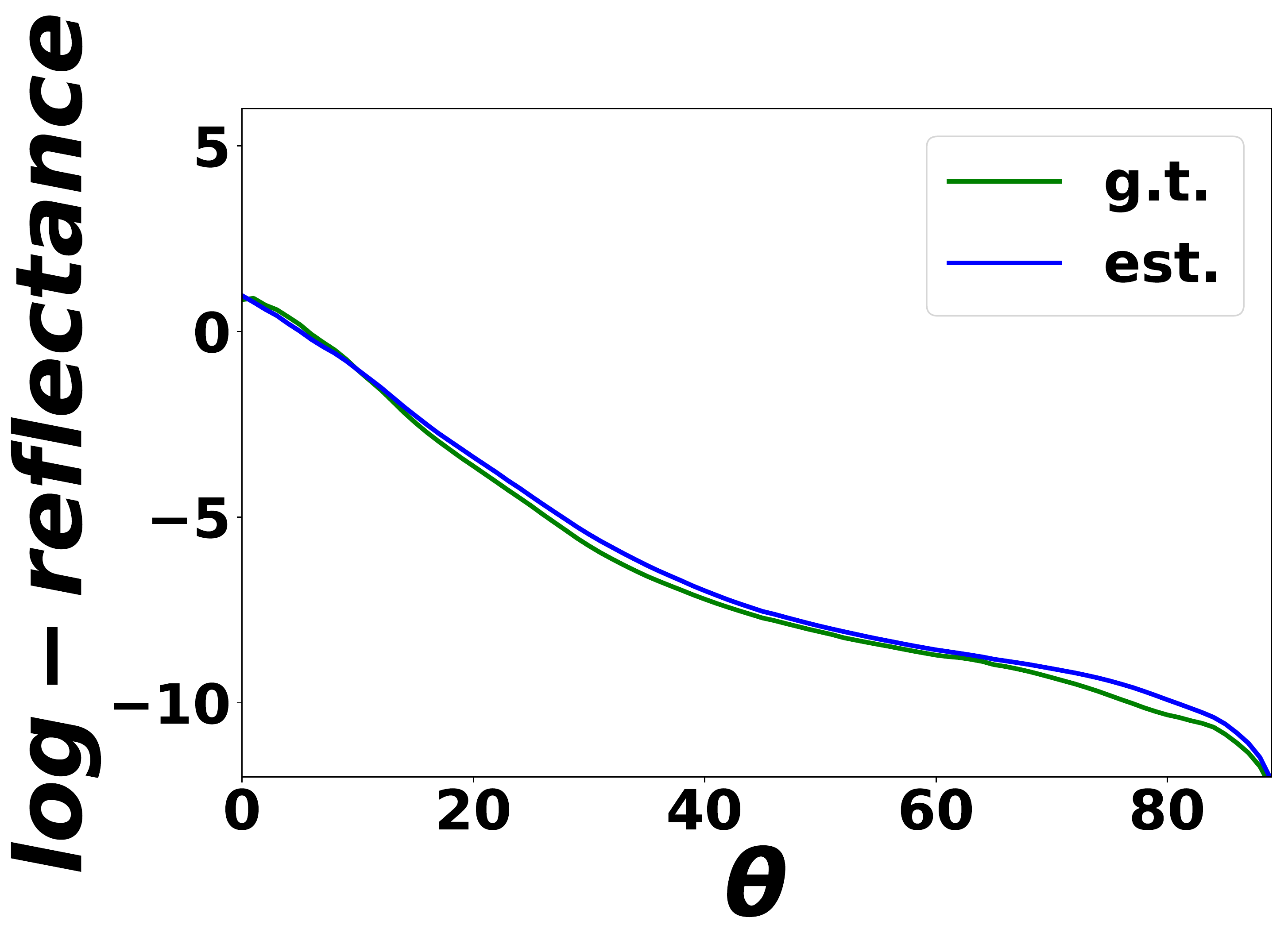}
         \caption{\footnotesize black-phenolic}
     \end{subfigure}
    \hspace*{-0.8cm}
    \begin{subfigure}[b]{0.18\textwidth}
         \centering
         \includegraphics[trim={70 0 10 0},clip, height=0.7\textwidth]{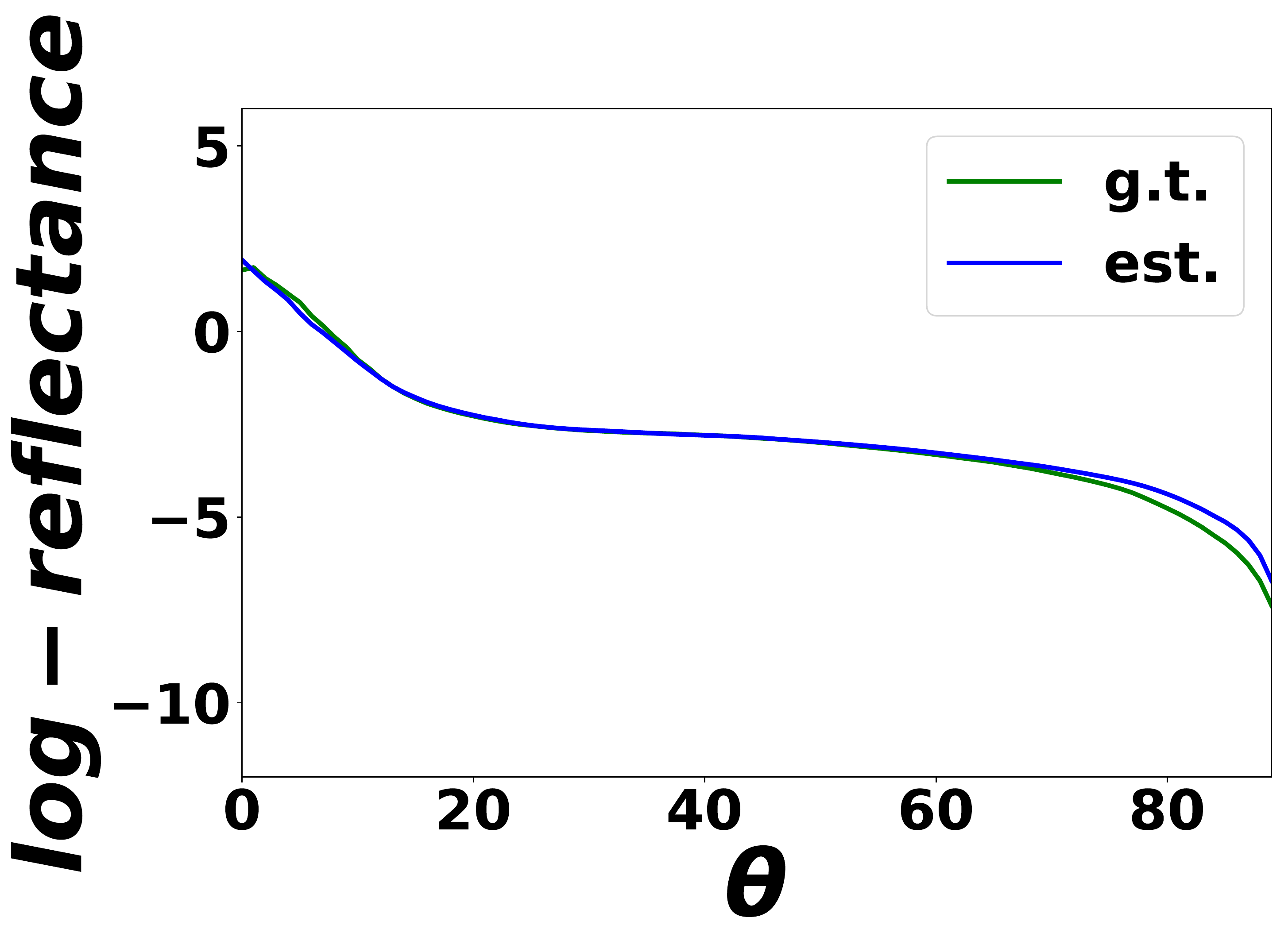}
         \caption{white-marble}
     \end{subfigure}
     \caption{\small Recovered BRDF curves versus the ground truth BRDFs for three different materials. }\label{fig:brdfrecovery}
\end{figure}

We also present examples of recovered (non-Lambertian) BRDFs compared with their corresponding ground-truths, as shown in Fig-\ref{fig:brdfrecovery}.   

\subsubsection{Convergence Analysis} 
As our energy model and minimization algorithm are mostly heuristic, it is difficult to prove the convergence on shape and BRDF metrics compared to ground truths, and such proof is beyond the scope of this paper. Instead, we offer to verify the convergence experimentally.

\begin{figure}[!htb]
    \centering
    \hspace*{-0.5cm}
    \begin{subfigure}[b]{0.18\textwidth}
         \centering
         \includegraphics[trim={0 0 10 0},clip,height=0.65\textwidth]{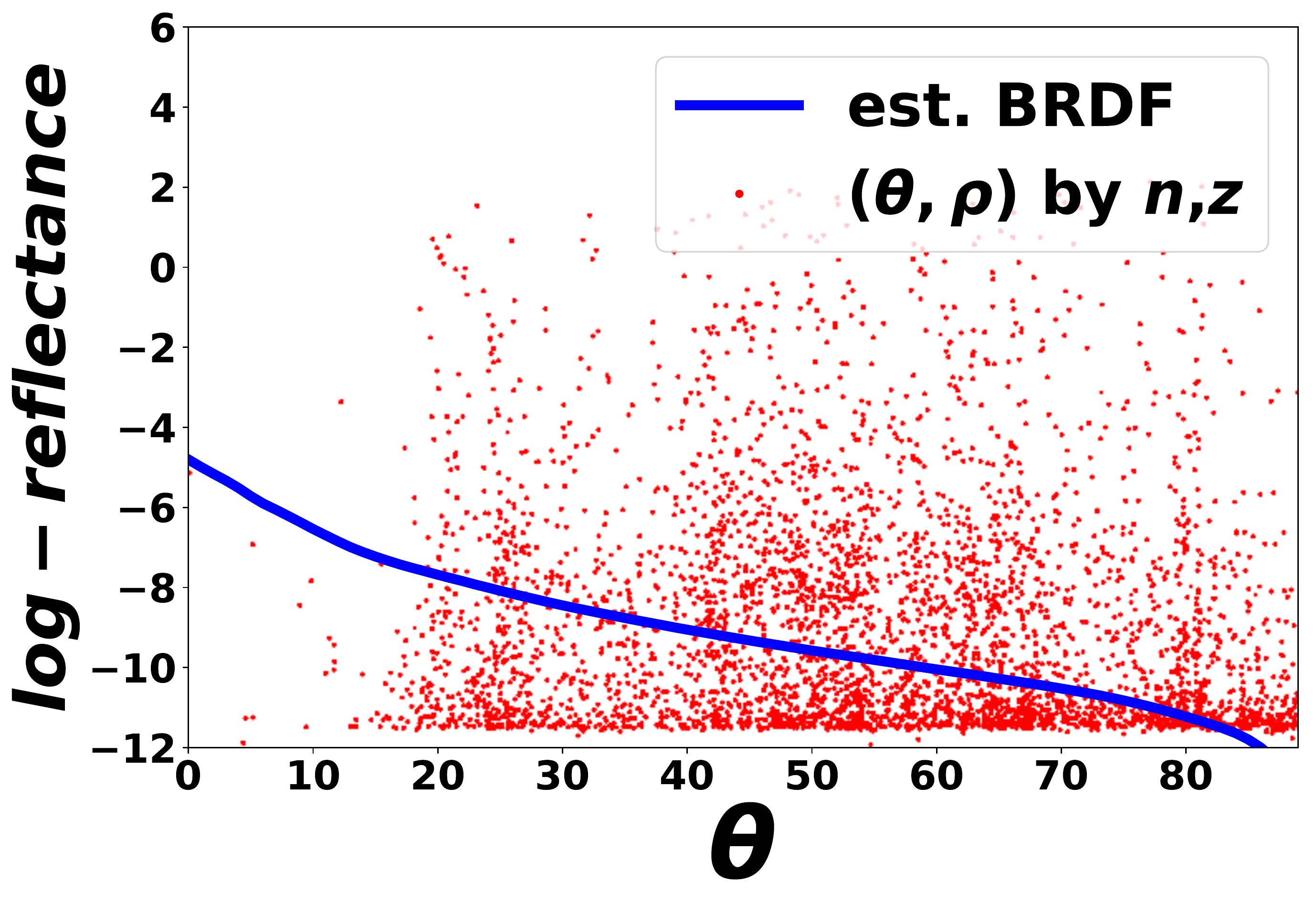}
         \caption{\footnotesize 1$_\text{st}$ iteration}
     \end{subfigure}
    \hspace*{-0.5cm}
    \begin{subfigure}[b]{0.18\textwidth}
         \centering
         \includegraphics[trim={60 0 10 0},clip,height=0.65\textwidth]{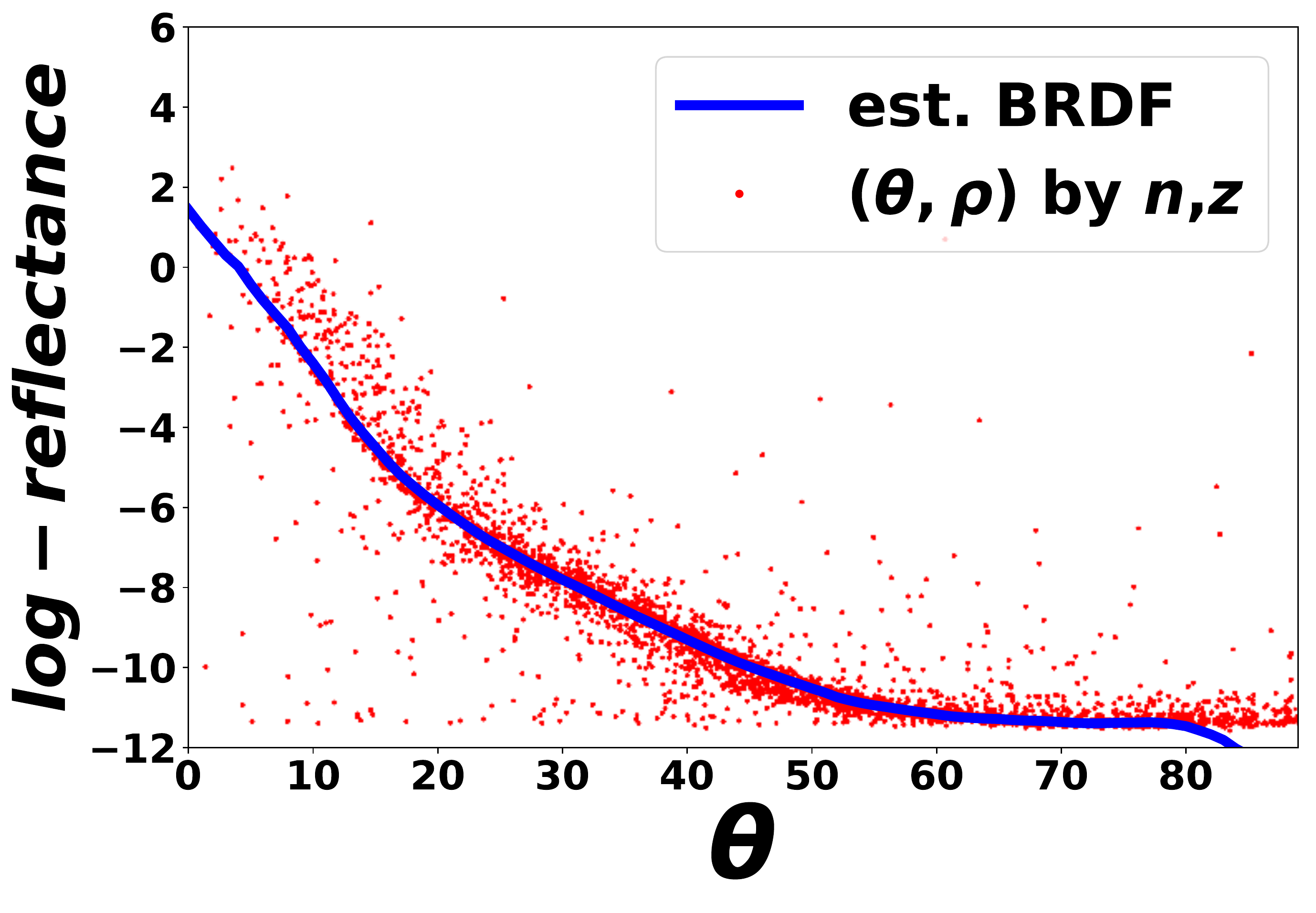}
         \caption{\footnotesize 10$_\text{th}$ iteration}
     \end{subfigure}
    \hspace*{-0.6cm}
     \begin{subfigure}[b]{0.18\textwidth}
         \centering
         \includegraphics[trim={60 0 10 0},clip,height=0.65\textwidth]{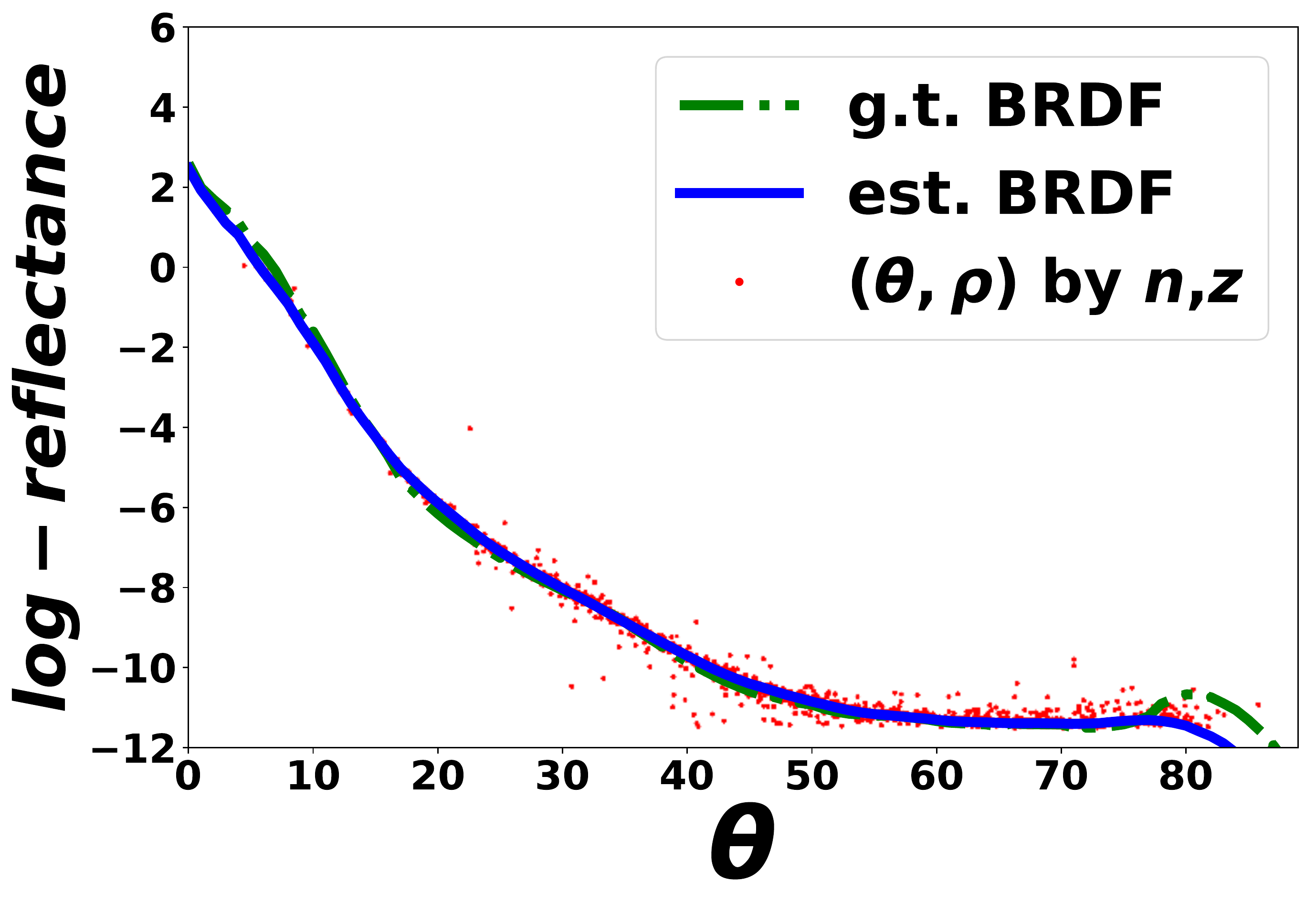}
         \caption{\footnotesize  50$_\text{th}$ iteration}\label{final}
     \end{subfigure}
    \caption{\small Visualization of how multi-view photometric constraints are gradually approached as the algorithm iterates. Red points are the reflectances retrieved from current shape, and the blue curve is the fitted BRDF curve. Both fittings gradually converge to ground-truth.} 
    \label{fig:visual_constraint}
\end{figure}

\begin{figure}[!htb]
\centering
         \includegraphics[height=0.15\textwidth,trim={2.2cm 0 0 0},clip]{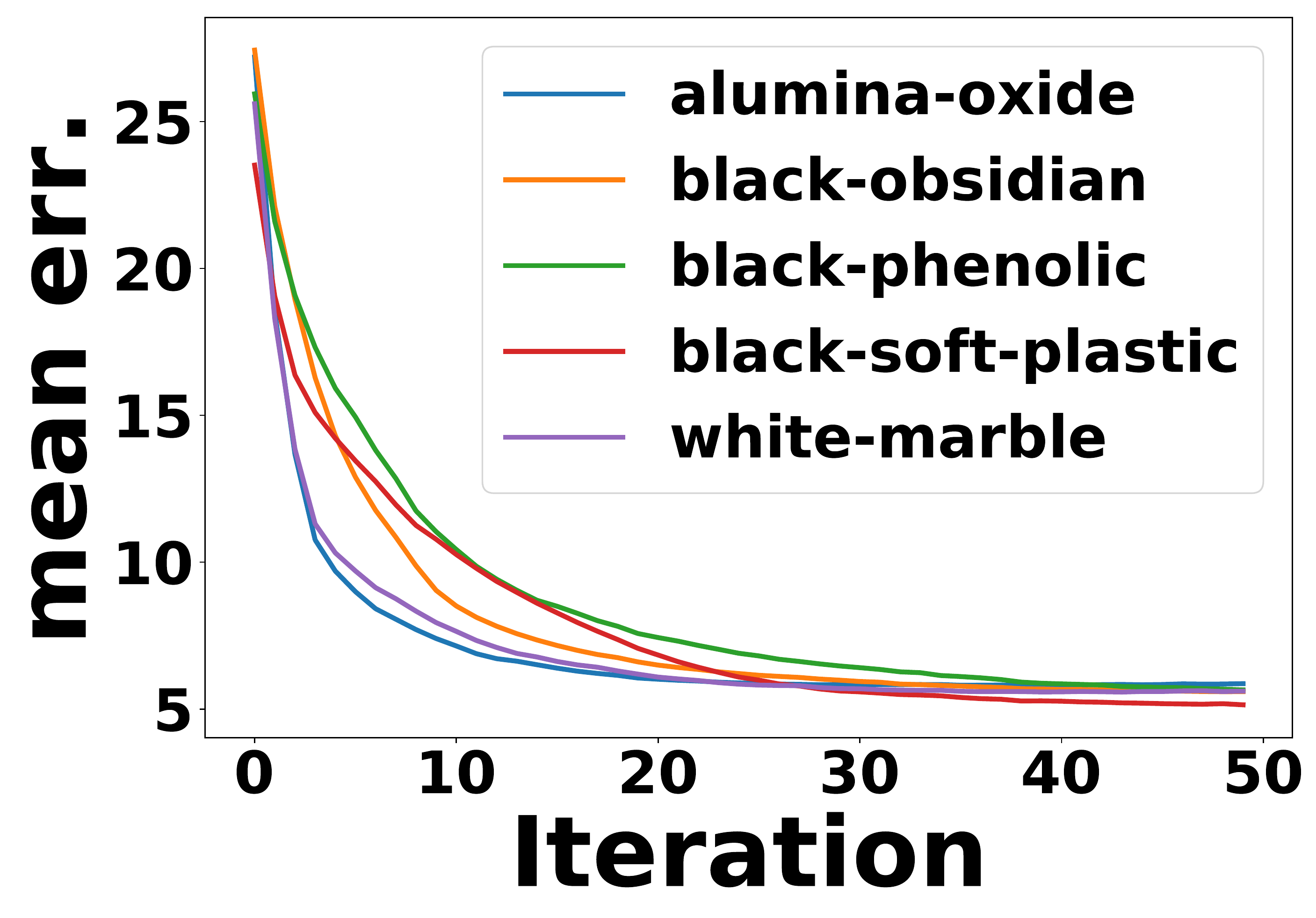}~~~
         \includegraphics[height=0.15\textwidth,trim={2.2cm 0 0 0},clip]{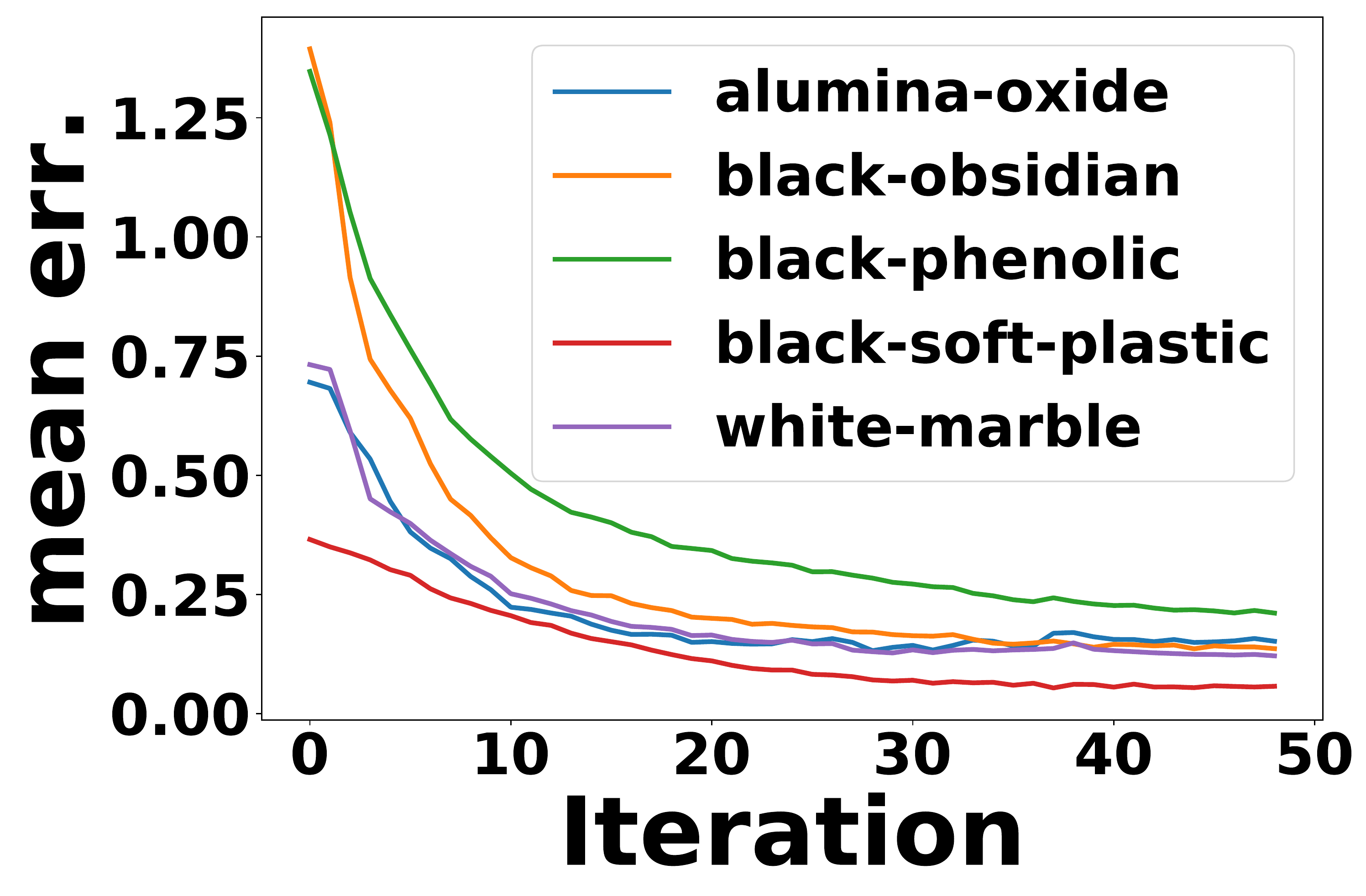}
     \caption{\small Left: normal angular error and Right: log-BRDF error (averaged over input angles) versus iterations.}\label{fig:errors_vs_iteration}
\end{figure}

Fig.~\ref{fig:visual_constraint} depicts how well the image formation equation (Eq.~\eqref{eq:formation}) is satisfied as the energy decreases. The vertical distances between each point and the predicted BRDF curve contribute to the overall inequality of multi-view photometric constraint Eq.~\eqref{eq:formation}. As iterations increase, this distance gradually decreases and the energy is minimized. The final solution in Fig.~\ref{final} is well-constrained and close to the true BRDF. Fig.~\ref{fig:errors_vs_iteration} illustrates error metrics indeed converge as a function of iterations.   

\begin{figure*}[!h] 
\centering
    \begin{subfigure}[b]{0.2\textwidth}
        \includegraphics[width=\textwidth,trim={22cm 22cm 20cm 25cm}, clip]{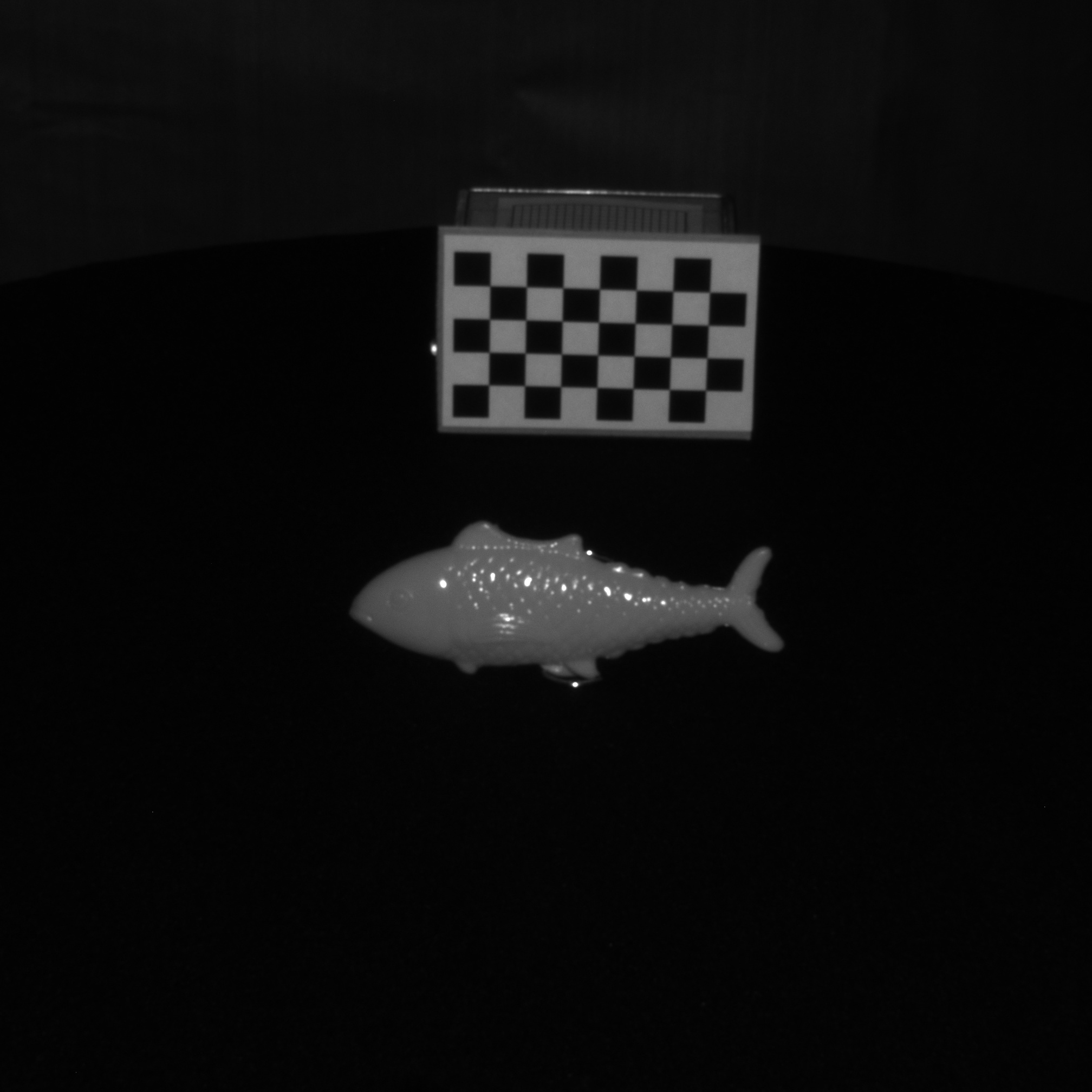}
        \includegraphics[width=\textwidth,trim={2cm 4cm 2cm 2cm}, clip]{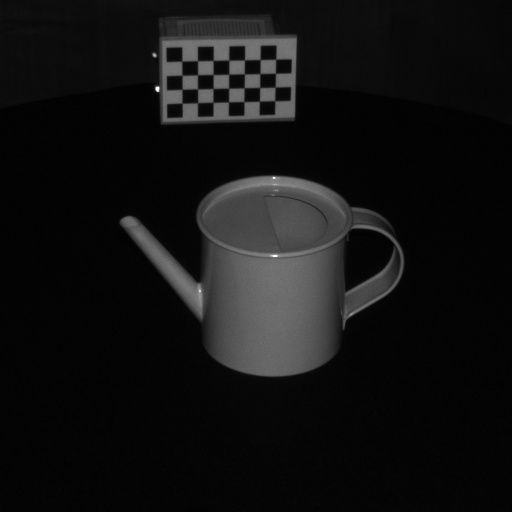}
        \includegraphics[width=\textwidth,trim={24cm 26cm 20cm 20cm}, clip]{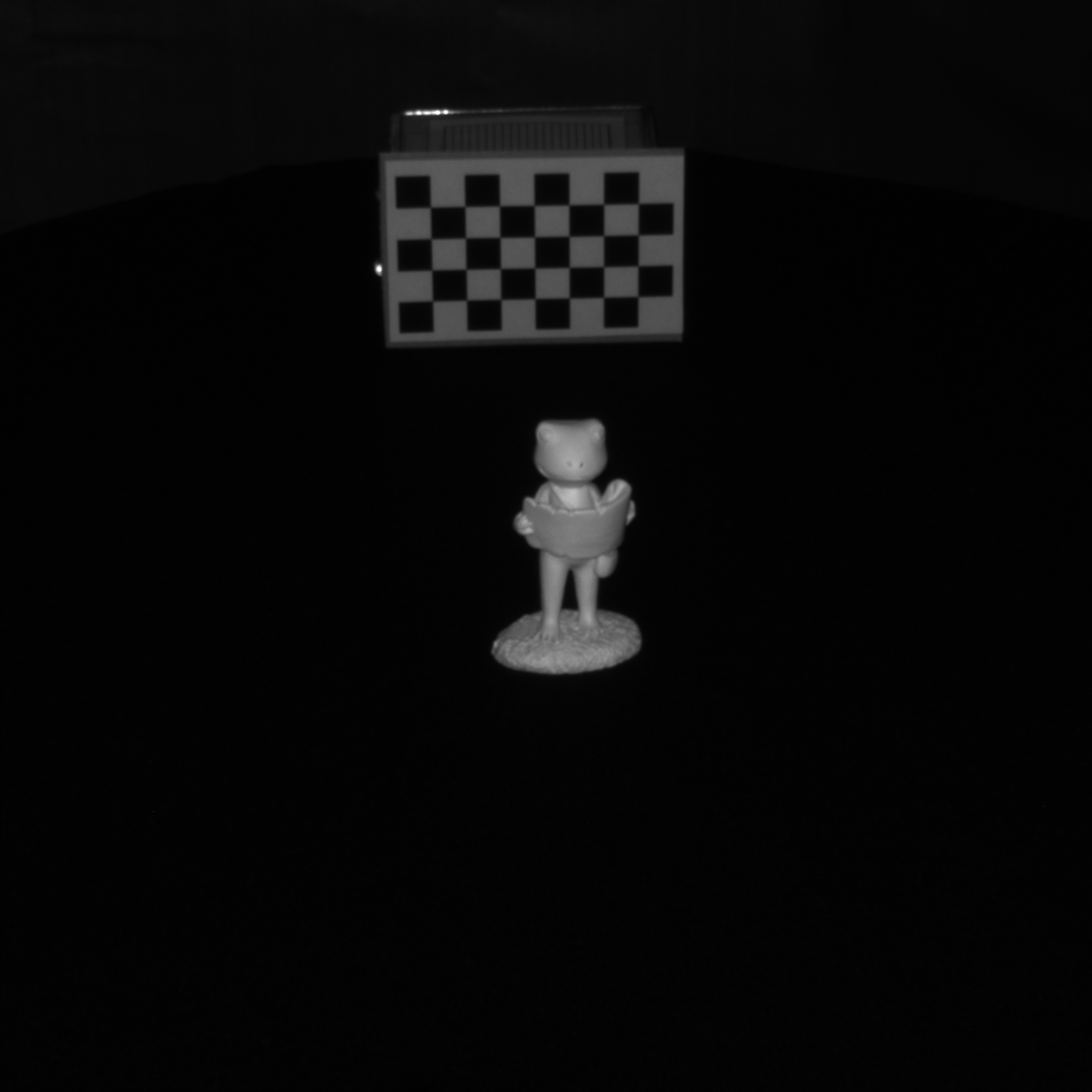}
        \caption{\footnotesize ref.view}
    \end{subfigure}
    \begin{subfigure}[b]{0.2\textwidth}
        \includegraphics[width=\textwidth,trim={11cm 11cm 10cm 12.5cm}, clip]{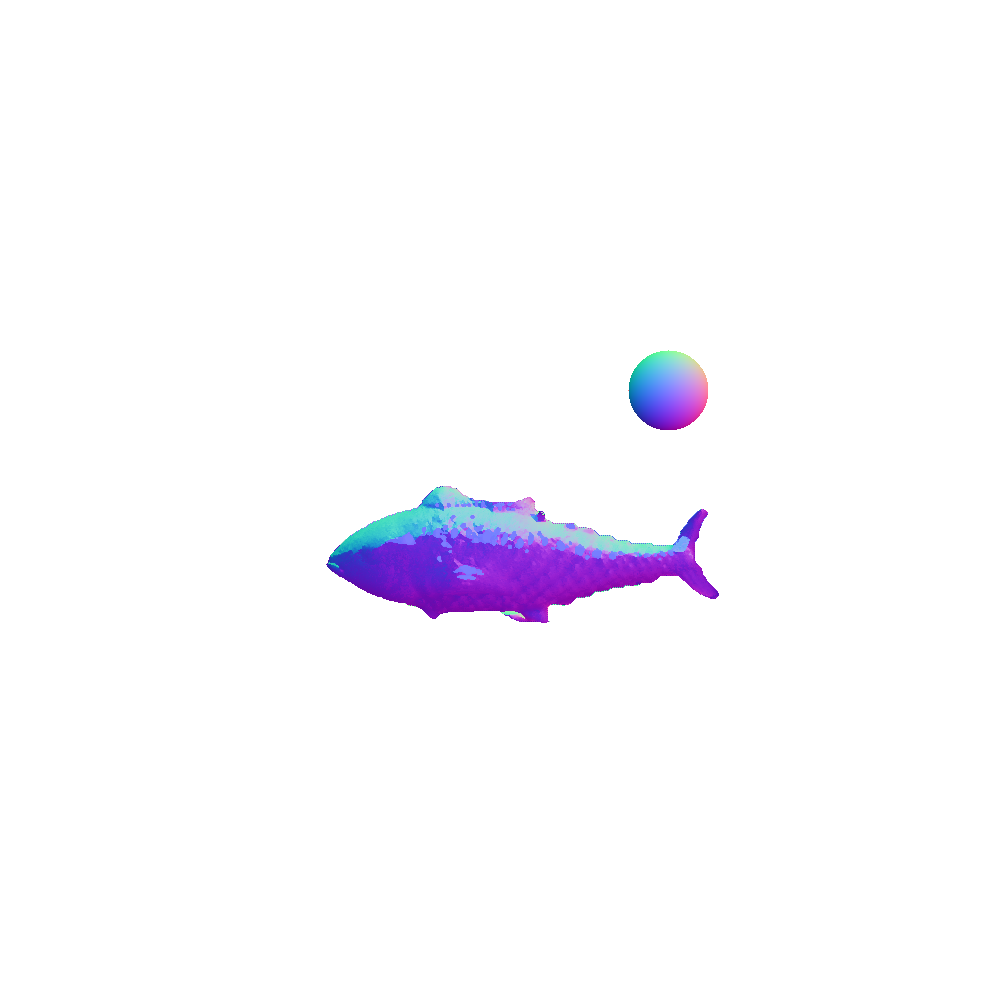}
        \includegraphics[width=\textwidth,trim={2cm 4cm 2cm 2cm}, clip]{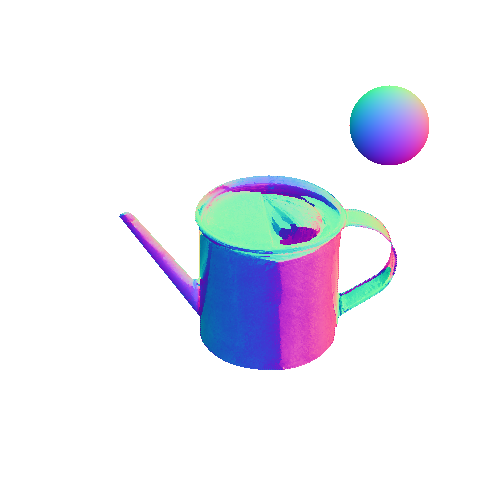}
        \includegraphics[width=\textwidth,trim={6cm 6.5cm 5cm 5cm}, clip]{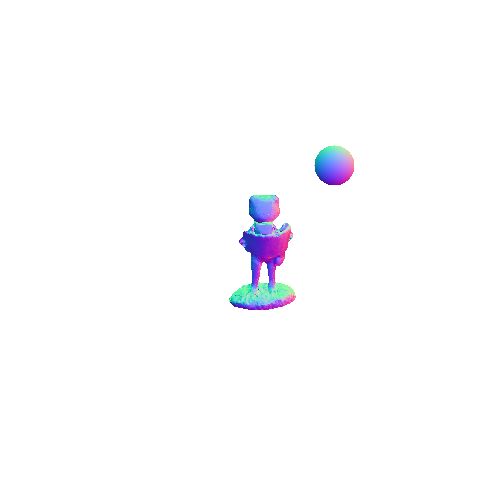}
        \caption{\footnotesize est. normal}
    \end{subfigure}
    \begin{subfigure}[b]{0.2\textwidth}
        \includegraphics[width=\textwidth,trim={11cm 11cm 10cm 12.5cm}, clip]{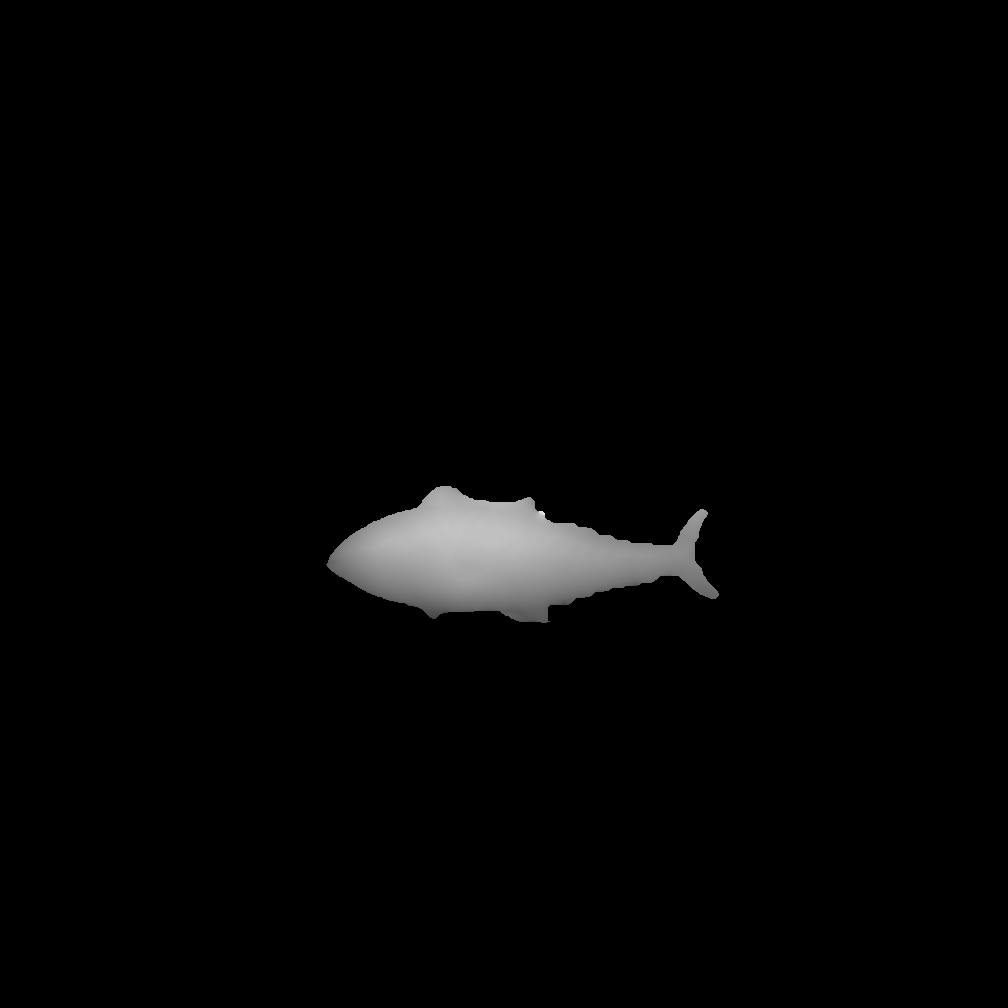}
        \includegraphics[width=\textwidth,trim={2cm 4cm 2cm 2cm}, clip]{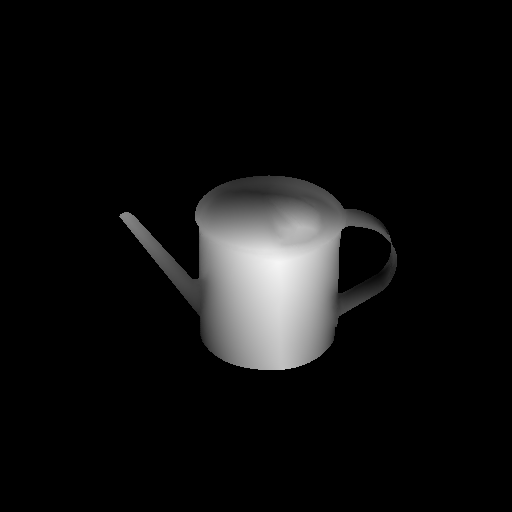}
        \includegraphics[width=\textwidth,,trim={6cm 6.5cm 5cm 5cm}, clip]{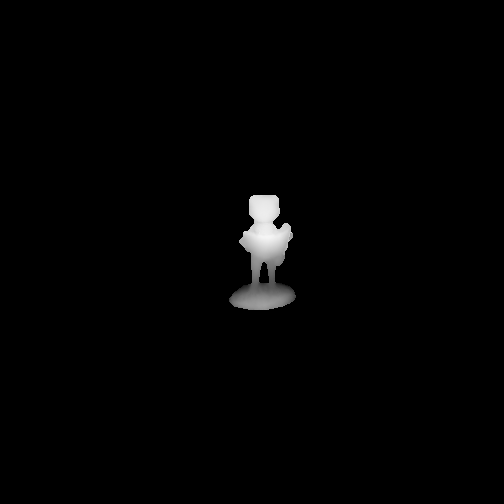}
        \caption{\footnotesize est. depth}
    \end{subfigure}
    \begin{subfigure}[b]{0.2\textwidth}
        \includegraphics[width=\textwidth,trim={11cm 11cm 10cm 12.5cm}, clip]{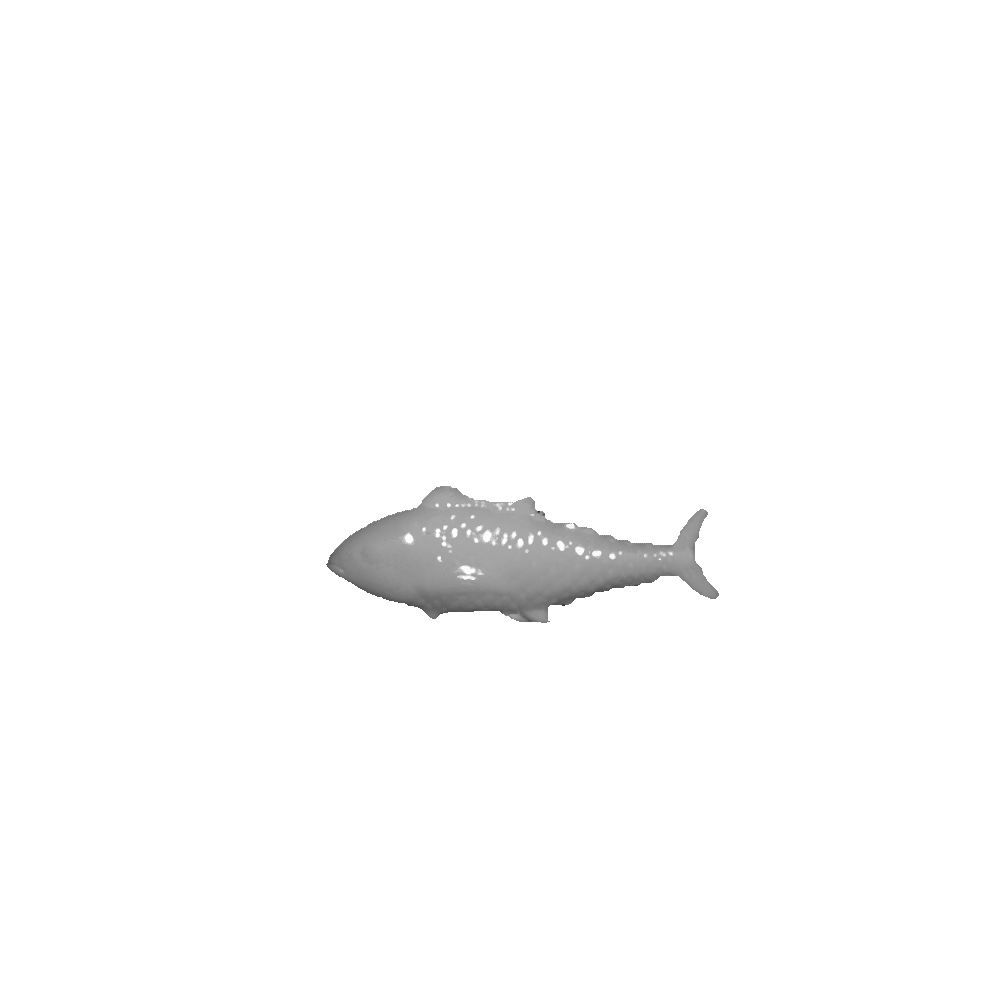}
        \includegraphics[width=\textwidth,trim={2cm 4cm 2cm 2cm}, clip]{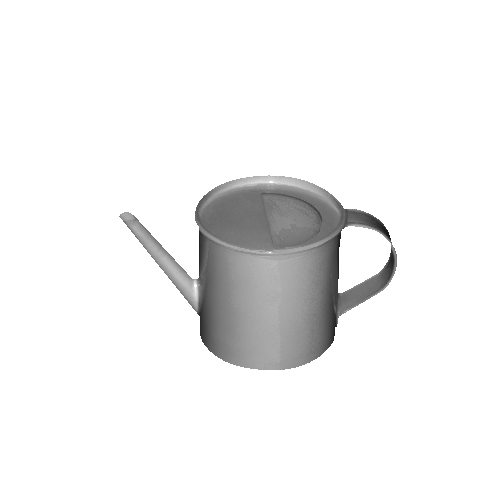}
        \includegraphics[width=\textwidth,,trim={6cm 6.5cm 5cm 5cm}, clip]{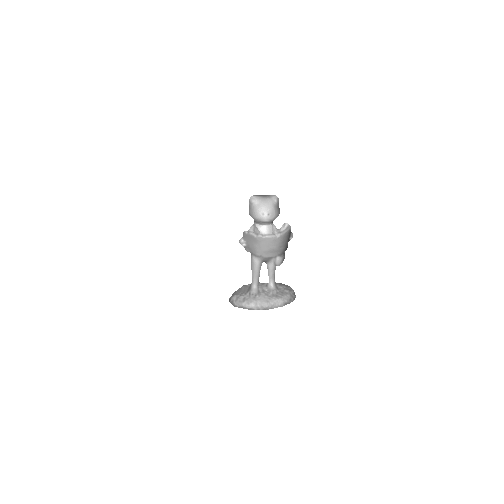}
        \caption{\footnotesize re-rendered}
    \end{subfigure}
   \caption{\small Input real image, recovered normal map, depth map, and re-rendered shape reconstruction.}\label{fig:reslimageresult} 
\end{figure*}

\begin{figure}[!htb] 
\centering
\begin{subfigure}[b]{0.2\textwidth}
        \includegraphics[width=\textwidth]{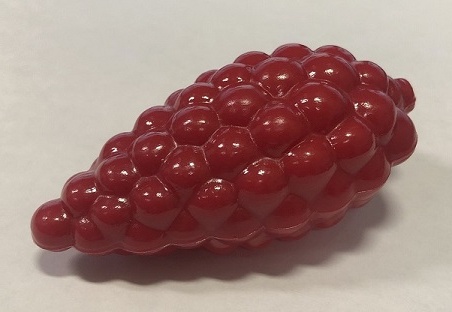}
        \includegraphics[width=\textwidth,trim={0cm 1.5cm 2cm 7cm}, clip]{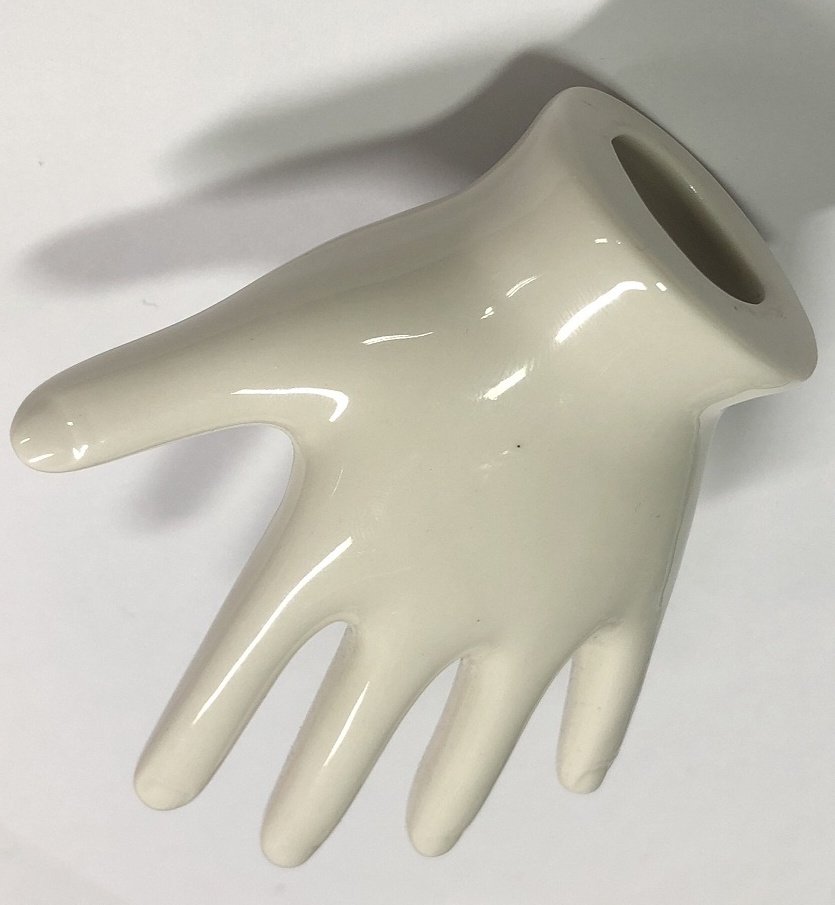}
        \caption{\footnotesize real shape}
    \end{subfigure}
    \begin{subfigure}[b]{0.2\textwidth}
        \includegraphics[width=\textwidth,trim={12cm 13cm 10cm 12cm}, clip]{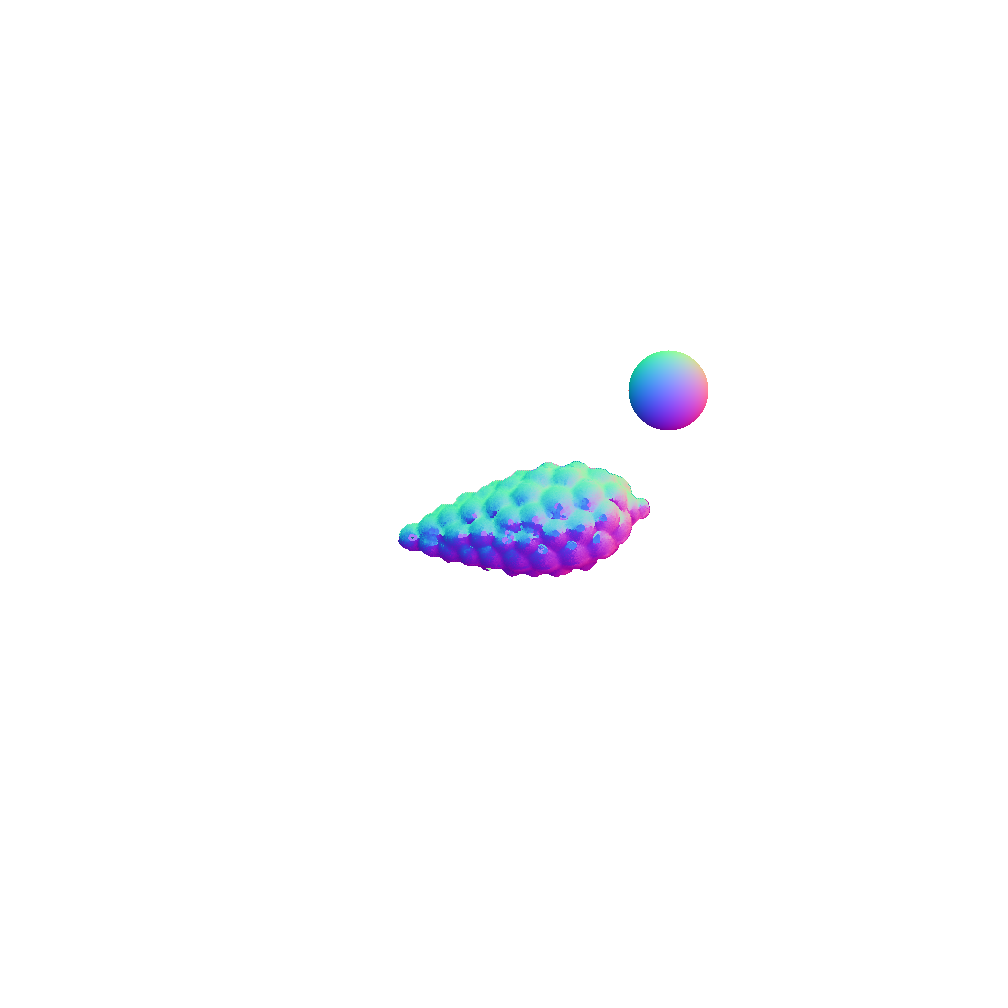}
        \includegraphics[width=\textwidth,trim={12cm 12cm 10cm 13cm}, clip]{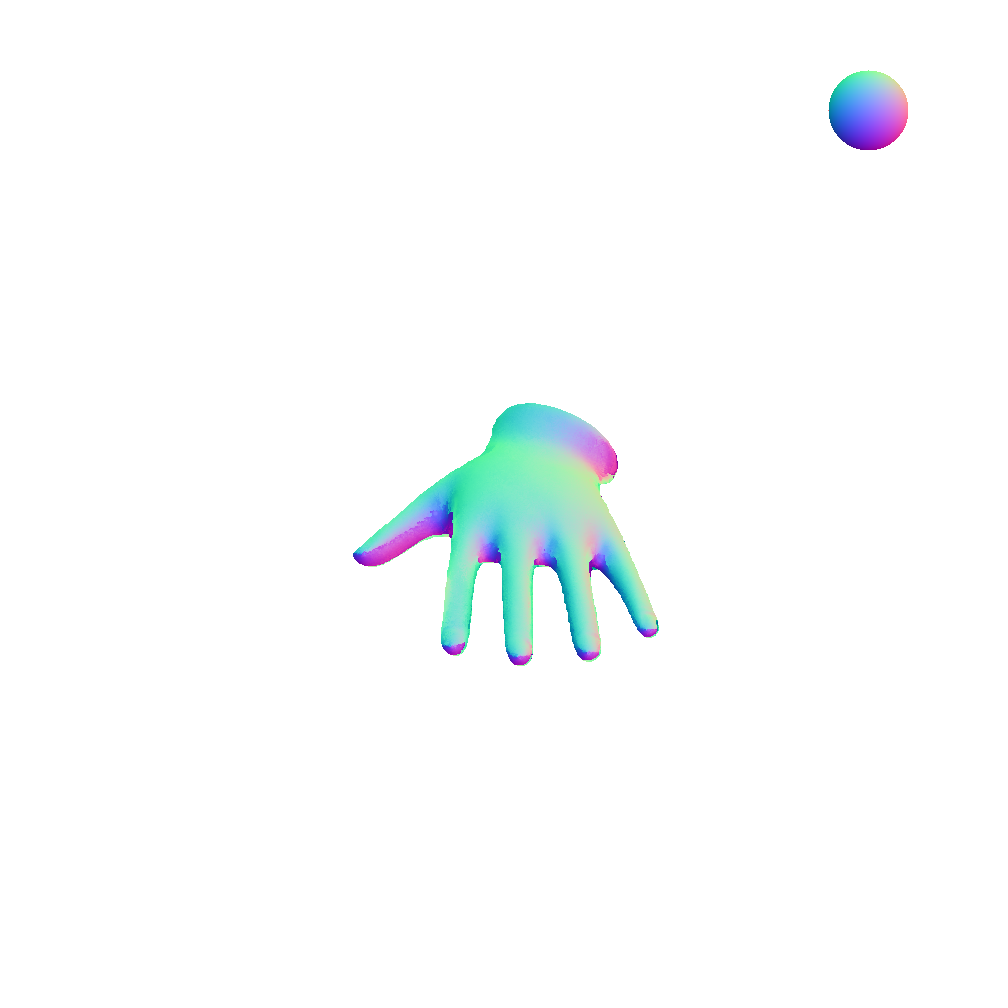}
        \caption{\footnotesize est. normal}
    \end{subfigure} \centering
\caption{\small Naturally lit real world objects versus reconstructed shapes.}\label{fig:realvsrecovered}
\vspace*{-2em}
\end{figure}

\subsection{Tests on real images}
In this subsection we qualitatively evaluate the performance of proposed algorithm on real-world images. To build the co-located camera and light source, we rigidly attached a universal light source onto the camera lens, as shown in Fig.~\ref{fig:device}. Similar to synthetic experiments, we take 20 images at varying camera positions in front of each target object. Since our method relies on photometric measurements, throughout our experiments, we use RAW image format and ensure that the measured image intensity is produced from a linear response function.  This is readily accessible for commodity DSLR cameras and many smartphone cameras.  

For obtaining extrinsic camera parameters, a standard checkerboard is placed near the object, and intrinsic and extrinsic parameters are solved by minimizing the reprojection error on corner points (cf. Fig.~\ref{fig:reslimageresult}) \cite{liu2017efficient,liu2017robust,wang2016slam}.  Input images are then rectified and foreground regions are manually cropped out.  More experimental results on real data are shown in Fig.~\ref{fig:reslimageresult} and \ref{fig:realvsrecovered}.  It is clear that our method achieves fine-grained reconstruction; even small details (such as the tiny surface bumps) are recovered vividly.

\section{Conclusion}
This paper has presented a new multi-view photometric 3D reconstruction method built upon a simple and practical camera-light configuration. It is able to recover fine-detailed 3D shape of a purely texture less surface with unknown arbitrary (non-Lambertian) reflectances, from a small set of multi-view input images. Our key contribution is a new optimization procedure that solves the challenging (highly non-convex) energy minimization task effective and optimally, without proper initialization.  Our method obtains visually compelling results on both synthetic data and real images.  Possible future extensions include to relax the assumptions about the scene, such as isotropic BRDF, uniform material, or point light source. The co-located configuration may also be relaxed. One limitation of our methods is that it can only handle open smooth (or piecewise smooth) surface visible by the reference camera view, and assume uniform reflectance on its surface. While we are working on relaxing these limitations, we hope this work may inspire future researchers working in the field.

\section{Acknowledgement}
This research is funded in part by the ARC Centre of Excellence for Robotics Vision (CE140100016), ARC-Discovery (DP 190102261), JSPS KAKENHI (JP20H05951) and by the Ministry of Education, Science, Sports and Culture Grant-in-Aid for Scientific Research on Innovative Areas (JP15H05918). We would like to thank Liu Liu for his support in camera calibration.

{\small
\bibliographystyle{ieee_fullname}
\bibliography{egbib}
}

\end{document}